\documentclass[paper,onecolumn,long,noendfloat,new,nonreprod,noblind]{geophysics}

\usepackage{url}
\usepackage{amsmath, amssymb}
\usepackage{longtable}
\usepackage{caption}
\usepackage[]{hyperref}
\hypersetup{
    pdftitle={Modeling IP with VAE},
    pdfauthor={Charles L. Bérubé},
    bookmarksnumbered=true,     
    bookmarksopen=true,         
    bookmarksopenlevel=1,       
    colorlinks=true,          
    allcolors=blue,
    pdfstartview=Fit,           
    pdfpagemode=UseOutlines,    
    pdfpagelayout=TwoPageRight
}
\usepackage{hypcap}

\newcommand{\lv}{\left\lvert}
\newcommand{\rv}{\right\rvert}
\newcommand\given[1][]{\,#1\lvert\,}

\begin{document}

\title{Data-driven modeling of time-domain induced polarization}

\renewcommand{\thefootnote}{\fnsymbol{footnote}} 

\author{Charles L. Bérubé\footnotemark[1] and Pierre Bérubé\footnotemark[2] \vspace{0.5em}}

\address{
\footnotemark[1]Polytechnique Montréal\\
Department of civil, geological and mining engineering\\
C.P. 6079, succ. Centre-ville, Montréal QC Canada H3C 3A7\\
\vspace{0.5em}\footnotemark[2]Abitibi Geophysics Inc.\\
1740 ch. Sullivan, Val-d'Or QC Canada J9P 7H1
\vspace{0.5em}
}

\footnotetext[1]{\textbf{email:} charles.berube@polymtl.ca\\}


\lefthead{Bérubé \& Bérubé}
\righthead{Data-driven IP modeling}

\renewcommand{\figdir}{Fig} 

\maketitle

\begin{abstract}
We present a novel approach for data-driven modeling of the time-domain induced polarization (IP) phenomenon using variational autoencoders (VAE). VAEs are Bayesian neural networks that aim to learn a latent statistical distribution to encode extensive data sets as lower dimension representations. We collected 1\,600\,319 IP decay curves in various regions of Canada, the United States and Kazakhstan, and compiled them to train a deep VAE. The proposed deep learning approach is strictly unsupervised and data-driven: it does not require manual processing or ground truth labeling of IP data. Moreover, our VAE approach avoids the pitfalls of IP parametrization with the empirical Cole-Cole and Debye decomposition models, simple power-law models, or other sophisticated mechanistic models. We demonstrate four applications of VAEs to model and process IP data: (1) representative synthetic data generation, (2) unsupervised Bayesian denoising and data uncertainty estimation, (3) quantitative evaluation of the signal-to-noise ratio, and (4) automated outlier detection. We also interpret the IP compilation's latent representation and reveal a strong correlation between its first dimension and the average chargeability of IP decays. Finally, we experiment with varying VAE latent space dimensions and demonstrate that a single real-valued scalar parameter contains sufficient information to encode our extensive IP data compilation. This new finding suggests that modeling time-domain IP data using mathematical models governed by more than one free parameter is ambiguous, whereas modeling only the average chargeability is justified. A pre-trained implementation of our model---readily applicable to new IP data from any geolocation---is available as open-source Python code for the applied geophysics community.
\end{abstract}

\newpage
\section{Introduction}

Mineral explorers have used the induced polarization (IP) method extensively since the 1950s \citep{seigel_early_2007}. The IP phenomenon refers to the temporary and reversible energy storage by the Earth's subsurface (or porous media in general) when subjected to a variable external electric field. This phenomenon is generally amplified by the presence of metallic particles in the subsurface \citep{revil_induced_2015, misra_interfacial_2016}, although experimental results have recently shown that the textural attributes of altered rocks can impede polarization, effectively passivating the metallic particles \citep[e.g.,][]{berube_mineralogical_2018, gurin_induced_2019}. Still, to this day, the IP method is one of the primary tools used by exploration geophysicists to characterize sulfide-associated ore systems, including volcanogenic massive sulfide \citep{tavakoli_deep_2016}, intrusion-related disseminated gold \citep{berube_mineralogical_2018}, porphyry copper \citep{close_electrical_2001} and iron oxide-copper-gold deposits \citep{aguilef_relationship_2017}. Since the early 2000s, researchers have also started leveraging the polarization of metallic particles for near-surface environmental applications. Examples include characterizing potential acid mine drainage generators such as mining slag heaps \citep{gunther_spectral_2016} and tailings \citep{placencia-gomez_spectral_2015}, monitoring high-pressure injection of iron particles used for groundwater remediation \citep{flores_orozco_monitoring_2015}, and detecting the precipitation of metal sulfides associated with bioremediation of uranium-contaminated groundwater \citep{flores_orozco_using_2011}. 

It is almost always necessary to process and clean IP data before proceeding to inverse modeling. However, processing IP data can prove to be a tedious, time-consuming, and somewhat subjective process. IP data processing is further complicated when the IP signal-to-noise ratio (S/N) is weak due to unfavorable field conditions (e.g., highly conductive overburden in wetlands or high electrode contact resistance in dry environments). While modeling techniques can quickly eliminate systematic noise sources \citep[e.g.,][]{olsson_doubling_2016}, IP processing steps still involve a careful visual inspection of every voltage decay curve collected during a survey. During this inspection, the interpreter can remove specific time windows considered outliers compared to their neighbors or even remove entire decay curves when they stray too far from what is considered ``standard'' (Aarhus Workbench demonstration, 4th International Workshop on IP, 2016). Therefore, two interpreters may not agree on what defines an outlier, what exactly is a standard decay curve, or what S/N is considered acceptable depending on specific field conditions. \cite{flores_orozco_decay_2018} introduced a rule-based approach to mitigate the subjectivity of visual IP decay curve inspection curves. However, their approach relies on fitting reference chargeability curves with an empirical power-law model assumption. More recently, \cite{barfod_automatic_2021} proposed to leverage machine learning to accelerate and standardize IP data processing. The main pitfall of their approach is that it relies on supervised learning, which by definition requires manual labeling of extensive training data sets. The first motivation for our study is to leverage unsupervised machine learning to process IP data in a data-driven manner, effectively removing the need for manual labeling and eliminating the need for simplified empirical models. We also define our method under a Bayesian framework which allows proper data uncertainty estimation. We further discuss the differences between our work and the approaches presented in \cite{olsson_doubling_2016}, \cite{flores_orozco_decay_2018}, and \cite{barfod_automatic_2021} in the Discussion section.

Inverse modeling usually follows the cleaning and processing of IP data. In its most basic form, inverse modeling assumes that chargeability only alters the effective conductivity of the subsurface \citep{oldenburg_inversion_1994}. Other approaches consist of leveraging the full-wave recording of voltage decay curves to perform inverse modeling with the Cole-Cole model \citep{fiandaca_resolving_2013}, which is entirely empirical. However, \cite{berube_bayesian_2017} and \cite{madsen_time-domain_2017} criticized the Cole-Cole model due to its inability to represent actual data accurately, its ambiguity in the frequency domain, and its strong parameter covariance. IP researchers have been exploring alternatives to the Cole-Cole model for many years, either with reparametrization \citep{fiandaca_re-parameterisations_2018}, replacement with the Debye and Warburg decomposition approaches \citep{nordsiek_new_2008, weigand_debye_2016, berube_bayesian_2017, berube_mineralogical_2018}, or by developing mechanistic IP models.

The past decade saw the publication of significant advancements on mechanistic models explaining the IP response of metallic particles in porous media \citep[e.g.,][ and respective follow-up parts]{revil_induced_2015, misra_interfacial_2016, bucker_electrochemical_2018}. These formulations provide a holistic approach to IP modeling, including factors such as the intrinsic properties of the subsurface interstitial electrolyte and metallic inclusion geometry, nature, and volumetric content. For example, the models proposed in \cite{revil_induced_2015, misra_interfacial_2016} assume that pyrite and other metallic minerals are semiconductors and neglect the effect of redox-active ions that may permeate the metal-electrolyte interface. The model presented in \cite{bucker_electrochemical_2018} extends the model of \cite{wong_electrochemical_1979} for perfectly conducting metallic spheres, including the effect of redox-active cations. Consequently, mechanistic models have many parameters that need to be constrained or optimized to interpret experimental data. Mechanistic models have the advantage of providing a physical basis to IP interpretation, yet IP practitioners seldom use them in actual applications due to their complexity and over-parametrization \citep[see the Discussion section in][ and references therein]{bucker_electrochemical_2018}. The second motivation for this study is to take a step back from empirical and mechanistic models and investigate the use of data-driven methods for IP modeling. Our intent is not to discourage the use of mechanistic models. Instead, we make the first step toward a new approach that avoids prior assumptions about the IP phenomenon to focus on representing IP data as we measure it in the field.

The objective of this study is twofold. First, we aim to automate, accelerate and standardize the IP data processing workflow using an unsupervised data-driven approach. Second, we aim to learn new insights on IP parametrization by learning the lower-dimension representations of time-domain IP data. To reach these goals, we developed a Bayesian neural network that consists of an inference model (encoder) and an observation model (decoder), forming a variational autoencoder (VAE). We hereafter refer to our model as the IP-VAE. We trained the IP-VAE to encode one of the most extensive compilation of field IP data reported in the geophysical literature, consisting of 1\,600\,319 decay curves measured between 2018 and 2021 in various regions of Canada, the United States, and Kazakhstan. We start by describing the time-domain IP measurement principle and introduce our training and testing data sets. Next, we provide the theoretical background for the IP-VAE, describe the neural network training procedure, and present the methods used to evaluate the model's performance. We then demonstrate four applications of the IP-VAE: generative IP modeling, Bayesian denoising, S/N analysis, and outlier detection. Finally, we analyze the latent representation of our extensive IP data compilation and discussion its implications for IP parametrization and inverse modeling.

\section{IP data compilation}
This section provides an overview of the time-domain IP measurement principle and defines the various data sets used throughout this study.

\subsection{Measurement principle}
The time-domain IP measurement principle is illustrated in Figure~\ref{fig:tdip} and summarized as follows. An electric current is injected in a polarizable subsurface using a pair of current electrodes, while the voltage is measured using a separate pair of potential electrodes. Typically, the electric current follows a square pulse with a predefined on-off duty cycle of 1 to 8 seconds. During the injection time, the current is constant, leading to the measurement of a primary voltage ($V_\textrm{p}$) between the potential electrodes. After turning off the current, the voltage drops abruptly to a value that corresponds to the secondary voltage ($V_\textrm{s}$). Exponential decay of the voltage follows the abrupt drop to $V_\textrm{s}$ during the relaxation time (refer to Figure~\ref{fig:tdip}). $V_\textrm{p}$ comprises the effect of current flowing in the subsurface and the charge built up due to the polarizable material that it contains, whereas $V_\textrm{s}$ only accounts for the contribution of the polarizable material \citep{mao_induced_2016}. Chargeability is the parameter that is most commonly used to describe the polarization intensity of geomaterials. Chargeability is not exactly an intrinsic petrophysical property but an integrating parameter that describes the dispersive properties of electrical resistivity when measured at varying frequencies. Chargeability was initially defined by \cite{seigel_mathematical_1959} as
\begin{equation}
    m = \frac{\rho_0 - \rho_\infty}{\rho_0},
\end{equation}
where $\rho_0$ is the subsurface direct current resistivity, and $\rho_\infty$ is the subsurface resistivity at infinitely high frequency. Under the time-domain definition of IP, chargeability is the ratio between $V_\textrm{s}$ and $V_\textrm{p}$ \citep{viezzoli_revised_2009}. It is not possible to measure $V_\textrm{s}$ in practice. Instead, integrating the voltage decay curve in several predefined time windows ($t_j$) yields the unnormalized chargeability values. Standard measurement protocols include normalization of the chargeability values by $V_\textrm{p}$ and by the length of the integration windows to express chargeability in mV/V \citep{dahlin_measuring_2002}. The average chargeability ($\bar{m}$) then reads
\begin{equation}
    \bar{m} = \frac{1}{d}\sum_j^d \left[\frac{1}{t_j V_\textrm{p}} \int_{t_j} V(t)\,\textrm{d}t \right],
\end{equation}
where $d$ is the number of integration windows (the dimensionality of IP decays).

\plot{tdip}{width=\textwidth}
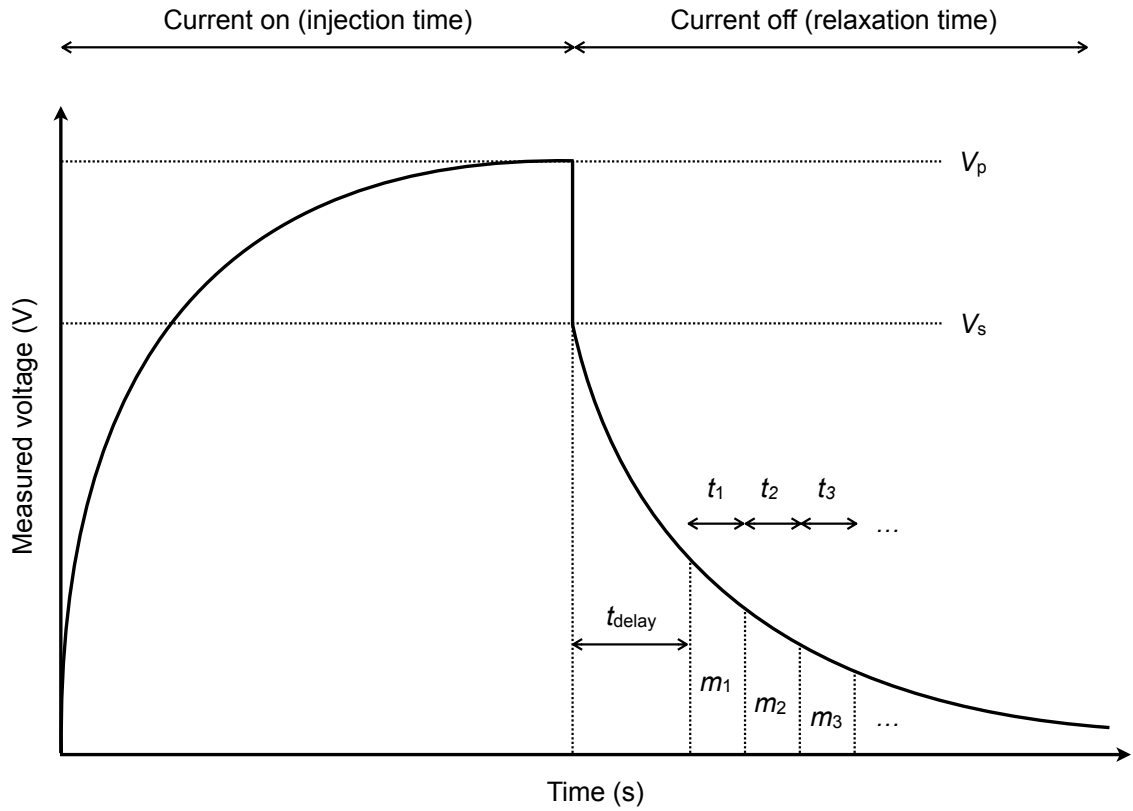
{Measurement principle of time-domain induced polarization. Chargeability values, denoted by $m_j$, are integrated in their respective predefined time windows ($t_j$) after a delay period ($t_\textrm{delay}$). Chargeability is normalized by the primary voltage ($V_\textrm{p}$) and integration time, and is therefore expressed in mV/V. All data reported in this study were obtained with $t_\textrm{delay} = 120$ ms and $t_j = 40$ ms.
}

\subsection{Training data}
We collected 1\,600\,319 IP decay curves by conducting 110 field surveys in Canada, the United States, and Kazakhstan and compiled them to train the IP-VAE. This training data set is hereafter denoted by $\mathcal{D}_{\textrm{Train}}$. Mineral exploration was the main purpose of all surveys in the compilation. $\mathcal{D}_{\textrm{Train}}$ comprises 47 surveys conducted in 2018, 25 surveys from 2019, 34 surveys from 2020, and 4 surveys from 2021. Appendix~A summarizes the basic parameters of each survey included in $\mathcal{D}_{\textrm{Train}}$ (Table~\ref{tbl:dtrain}). The surveys sites were diverse and included open fields, the Canadian boreal forest, glades, wetlands, outcrops, and more. The subsurface lithologies varied from granodiorite stock within the Canadian Shield (survey 20NT004, Wawa Gold Camp, Ontario) to serpentinite and amphibolite (survey 20FU072, Kazakhstan), for example. Approximately 85\% of the surveys in $\mathcal{D}_{\textrm{Train}}$ used a pole-dipole electrode array, 10\% a dipole-dipole array and 5\% a gradient array. 

We obtained all chargeability measurements with the IRIS ELREC Pro receiver. The IRIS TIPIX 2200 and 3000 transmitters, following a square pulse with an on-off duration of 1000 ms, served as the electrical current injection source for all surveys compiled in $\mathcal{D}_{\textrm{Train}}$. Integration of the voltage decay curve in 20 windows---as programmed in the ELREC Pro receiver's arithmetic mode---yielded chargeability values. The delay between the start of the relaxation time and the first integration window was 120 ms, and the duration of each integration window was 40 ms (refer to Figure~\ref{fig:tdip}).

\subsection{Synthetic test data}
After training, the IP-VAE was used to generate a synthetic data set consisting of 1.6M IP curves representative of the training data set. This data set, hereafter denoted by $\mathcal{D}_{\textrm{STest}}$, was used exclusively for testing purposes and comparing the denoising properties of the IP-VAE with other baseline methods. More details about the synthetic data generation process are in the Results section.

\subsection{Field test data}
We designed a field experiment to collect actual data at various S/N and test the denoising capabilities of the IP-VAE. First, we set up a dipole-dipole array configuration (dipoles 1 to 20) in a region of non-zero subsurface chargeability. Then, we injected the highest electrical current values allowed by the transmitter and field conditions (values ranged from 1.0 to 2.1 A). Doing so provided a first, high S/N version of the IP decays. Next, we divided the injected current by half by reducing the applied voltage at the transmitter and repeated the chargeability measurements without displacing the electrode array. By repeating these steps until no significant electrical current was injectable into the ground, we obtained 5 to 8 versions of the same IP decays with progressively lower S/N. Injected electrical current values for the low S/N versions ranged from 0.05 to 0.07 A. Finally, we repeated the experiment at 80 locations by moving the electrode array along a survey line and thus obtained a field test data set comprising several hundreds of high and low S/N examples of the same IP decays. Note that we stacked eight readings for each IP decay and performed the tests in under approximately 20 minutes at each location to mitigate the effects of telluric noise drift. The field test data set is hereafter denoted by $\mathcal{D}_{\textrm{FTest}}$.

\section{Methods}

\subsection{Variational autoencoder}
This section describes basic autoencoders (AE), provides an overview of the Bayesian framework that justifies VAEs, and presents the IP-VAE architecture and training process.

\subsubsection{Basic autoencoder}
AEs are neural networks designed to perform unsupervised data dimensionality reduction \citep{hinton_reducing_2006}. Common applications of AEs include denoising \citep{vincent_extracting_2008}, feature extraction \citep{meng_relational_2017}, image compression \citep{cheng_deep_2018} and outlier detection \citep{kieu_outlier_2019}. AEs aim to learn a mapping from the original input data to a low-dimensional (compressed) representation while simultaneously learning a mapping to reconstruct the original data using its compressed representation. An AE consists of two neural networks working in tandem: an encoder and a decoder.

A first set of internal trainable parameters $\phi$ governs the encoder (i.e., $\phi$ represents the weight matrices and bias vectors of linear encoding layers). The encoder is denoted by $g_{\phi}$ and its role is to map the original, high-dimensional data to one of its representations in a lower-dimensional space, hereafter referred to as the latent space. We denote vectors in the input data space by $\mathbf{x}$ and vectors in the latent space by $\mathbf{z}$. A second set of trainable parameters $\theta$ governs the decoder, which is denoted by $f_{\theta}$. $f_{\theta}$ aims to map $\mathbf{z}$ to the best possible reconstruction of $\mathbf{x}$. We denote reconstructed vectors in the output data space by $\mathbf{x'}$. The latent representations can therefore be expressed as $\mathbf{z} = g_{\phi}(\mathbf{x})$ and the reconstructed data can be expressed as $\mathbf{x'} = f_{\theta}(\mathbf{z})$. Combining these relationships the general equation for a basic AE reads
\begin{equation}
    \mathbf{x'} = f_{\theta}(g_{\phi}(\mathbf{x})),
\end{equation}
where $\phi$ and $\theta$ need to be optimized jointly. 

Considering a training data set of length $n$, denoted by $\mathcal{D}_\textrm{Train} = [\mathbf{x}^{(1)}, \mathbf{x}^{(2)}, \dots, \mathbf{x}^{(n)}]$, finding optimal values of $\phi$ and $\theta$ implies that the AE yields $\mathbf{x'}^{(i)}$ values that are close to $\mathbf{x}^{(i)}$. Optimization of deep neural networks is typically achieved with a variant of mini-batch gradient descent, whereby $\phi$ and $\theta$ are updated iteratively using backward propagation of errors \citep{goodfellow_deep_2016}. In the case of real-valued $\mathbf{x}^{(i)}$, such as IP decays, the preferred AE loss function ($\mathcal{L_\textrm{AE}}$) is the mean square error
\begin{equation}
    \mathcal{L}_{\textrm{AE}}(\phi, \theta, \mathbf{x}^{(i)}) = \left\lVert \mathbf{x}^{(i)} - \mathbf{x'}^{(i)} \right\rVert^2_2,
\end{equation}
which is computed and averaged over every mini-batch to update values of $\phi$ and $\theta$. 

\subsubsection{Bayesian framework}
Broadly speaking, VAE are probabilistic AE. VAEs aim to learn a generative distribution from which latent vectors can be sampled and decoded, ideally reconstructing the input data \citep{kingma_auto-encoding_2014}. Thus, the goal of training a VAE is to estimate the posterior distribution of the latent vectors $p_{\theta}({\mathbf{z}\given \mathbf{x}})$, where only the observations $\mathbf{x}$ are known. Intuitively, $p_{\theta}({\mathbf{z}\given \mathbf{x}})$ is a distribution that yields $\mathbf{z}$ values that are likely to generate $\mathbf{x}$ from the known data distribution. According to Bayes' theorem, the relationship between $\mathbf{x}$ and $\mathbf{z}$ reads
\begin{equation}
    p_{\theta}({\mathbf{z}\given \mathbf{x}}) = \frac{{p_{\theta}( {\mathbf{x}\given \mathbf{z}})\, p_{\theta}(\mathbf{z})}}{{p_{\theta}(\mathbf{x})}},
\label{eq:bayes}
\end{equation}
where $p_{\theta}(\mathbf{z})$ is a prior distribution for the latent vectors and $p_{\theta}( {\mathbf{x}\given \mathbf{z}})$ is the data likelihood. The denominator in equation~\ref{eq:bayes} (the evidence, or marginal likelihood) is defined by 
\begin{equation}
    p_\theta(\mathbf{x}) = \int p_\theta(\mathbf{x}\given\mathbf{z})\,p_\theta(\mathbf{z})\,d\mathbf{z},
\end{equation}
which is an intractable distribution because the search space for $\mathbf{z}$ is combinatorially large for complicated models (e.g., a neural network with hidden layers). The true $p_{\theta}({\mathbf{z}\given \mathbf{x}})$ is consequently also intractable \citep{kingma_auto-encoding_2014}. A tractable distribution $q_{\phi}({\mathbf{z}\given \mathbf{x}})$, also known as the inference model, is introduced to approximate $p_{\theta}({\mathbf{z}\given \mathbf{x}})$. This approximation is the basis for variational inference \citep{blei_variational_2017}, whereby the Kullback-Leibler divergence ($D_\textrm{KL}$) is used as a dissimilarity function to minimize the difference between $q_{\phi}({\mathbf{z}\given \mathbf{x}})$ and $p_{\theta}({\mathbf{z}\given \mathbf{x}})$. The objective of variational inference is therefore to minimize $D_\textrm{KL}\left[ {q_{\phi}\left( {\mathbf{z}\given\mathbf{x}} \right) \parallel p_{\theta}\left( {\mathbf{z}\given\mathbf{x}} \right)} \right]$,
which is used to derive the evidence lower bound \citep[ELBO,][]{kingma_auto-encoding_2014, blei_variational_2017}:
\begin{equation}
    \textrm{ELBO} = \mathbb{E}\log p_{\theta}\left( \mathbf{x}\given\mathbf{z} \right) - D_\textrm{KL}\left[ {q_{\phi}\left( \mathbf{z}\given\mathbf{x} \right)\parallel p_{\theta}\left( \mathbf{z} \right)} \right].
\label{eq:elbo}
\end{equation}
The first term on the right-hand side of equation~\ref{eq:elbo} is the data log-likelihood expectation. High values of the data log-likelihood indicate that the autoencoder is accurately reconstructing inputs using the latent vectors. The second term on the right-hand side of equation~\ref{eq:elbo} acts as a regularization term: it encourages $q_{\phi}\left( \mathbf{z}\given\mathbf{x} \right)$ to remain close to the latent distribution prior $p_{\theta}\left( \mathbf{z} \right)$. A common choice for $p_{\phi}(\mathbf{z})$---and by approximation $q_{\phi}({\mathbf{z}\given \mathbf{x}})$---is an isotropic normal distribution $\mathcal{N}[\boldsymbol{\mu}(\mathbf{x}), \boldsymbol{\sigma}(\mathbf{x})]$, whose mean ($\boldsymbol{\mu}$) and standard deviation ($\boldsymbol{\sigma}$) are vectors with the same dimensions as $\mathbf{z}$ \citep{kingma_auto-encoding_2014}. $q_{\phi}({\mathbf{z}\given \mathbf{x}})$ is effectively a probabilistic encoder because it depends on $\mathbf{x}$ to yield the mean and standard deviation of a normal distribution from which we can sample $\mathbf{z}$. 

The likelihood term in equation~\ref{eq:bayes} is the observation model, representing the probability of drawing examples from the data distribution given a latent vector. $p_{\theta}( {\mathbf{x}\given \mathbf{z}})$ is effectively a probabilistic decoder. Perhaps the most interesting benefit of using a probabilistic decoder over a deterministic one is that the former can act as a generative model \citep[e.g.,][]{nash_shape_2017, lim_molecular_2018}. In fact, for optimized values of $\phi$ and $\theta$, one can generate synthetic $\mathbf{x}$ values that are representative of $\mathcal{D}_{\textrm{Train}}$ by drawing values of $\mathbf{z} \sim \mathcal{N}(0, \mathbf{I})$ and decoding them with $p_{\theta}( {\mathbf{x}\given \mathbf{z}})$. The loss function used to optimize $\phi$ and $\theta$ is described next.

\subsubsection{VAE loss}
\cite{kingma_auto-encoding_2014} introduced a reparametrization trick to ensure that the probabilistic autoencoder loss is differentiable with respect to $\phi$ and $\theta$, allowing to train VAEs with gradient descent. The reparametrization trick consists of substituting $\mathbf{z} \sim q_{\phi}({\mathbf{z}\given \mathbf{x}})$, a random variable, by a deterministic variable $\mathbf{z} = \boldsymbol{\mu}(\mathbf{x}) + \boldsymbol{\epsilon}\odot\boldsymbol{\sigma}(\mathbf{x})$, where $\boldsymbol{\epsilon} \sim \mathcal{N}(0, \mathbf{I})$ is an auxiliary random vector with the same dimensions as $\mathbf{z}$. The $\odot$ symbol denotes the Hadamard (element-wise) product and $\mathbf{I}$ is the identity matrix.

The VAE loss function is defined as the negative ELBO and contains two terms \citep{kingma_auto-encoding_2014}. The first term is the Kullback–Leibler divergence ($D_{\textrm{KL}}$) between $q_{\phi}({\mathbf{z}\given \mathbf{x}})$, the approximate posterior, and $p_{\theta}(\mathbf{z})$, the prior for the latent vectors. The second term is $-\log{p_{\theta}({\mathbf{x}\given \mathbf{z}})}$, the negative log-likelihood (NLL) of generating real data with the decoder. The latter is a reconstruction loss akin to the one used in basic AE, because the parameters that maximize the log-likelihood in a Gaussian model are the same that minimize the mean square error \citep[see Chapter 5 of][]{goodfellow_deep_2016}. We assumed a standard normal distribution for $p_{\theta}(\mathbf{z})$, therefore the VAE loss function to be minimized by mini-batch gradient descent is \citep{kingma_auto-encoding_2014}
\begin{equation}
\begin{split}
    \mathcal{L}_{\textrm{VAE}}(\phi, \theta, \mathbf{x}^{(i)}) & = D_{KL} - \log{p_{\theta}({\mathbf{x}^{(i)}\given \mathbf{z}^{(i)}})} \\
    & = -\frac{1}{2} \sum_{k=1}^K \left(1 + \log{\sigma_k^2} - \mu_k^2 - \sigma_k^2 \right) + \left\lVert \mathbf{x}^{(i)} - \mathbf{x'}^{(i)} \right\rVert^2_2,  
\end{split}
\label{eq:vaeloss}
\end{equation}
where $K$ is the dimensionality of $\mathbf{z}$.

\subsubsection{Network architecture and training}
We implemented the IP-VAE within the PyTorch deep learning framework \citep{paszke_pytorch_2019}. Figure~\ref{fig:architecture} illustrates the network architecture of the IP-VAE. The encoder part of the network consists of an input layer, which accepts mini-batches of $\mathbf{x}$, followed by two fully connected hidden linear layers ($h^{(1)}$ and $h^{(2)}$). In the probabilistic encoder bottleneck, the $h^{(\mu)}$ layer yields $\boldsymbol{\mu}(\mathbf{x})$ while $h^{(\sigma)}$ simultaneously yields $\boldsymbol{\sigma}(\mathbf{x})$. These values are then used in the reparametrization trick to draw $\mathbf{z} = (z_1, z_2)$, a two-dimensional latent vector sample.

The decoder part of the network is nearly symmetrical to the encoder. Thus, samples of $\mathbf{z}$ go through two fully connected linear layers ($h^{(3)}$ and $h^{(4)}$), and a final output layer ($h^{(5)}$) returns the reconstructed data $\mathbf{x'}$. 

We experimented with various activation functions that introduce non-linearity between the various linear layers of the neural network, including the hyperbolic tangent, sigmoid, and rectified linear unit functions \citep[see Chapter 6 of][]{goodfellow_deep_2016}. The hyperbolic tangent function consistently provided the best reconstructions and was used as the nonlinear activation function where indicated in Figure~\ref{fig:architecture}.

\plot{architecture}{width=\textwidth}
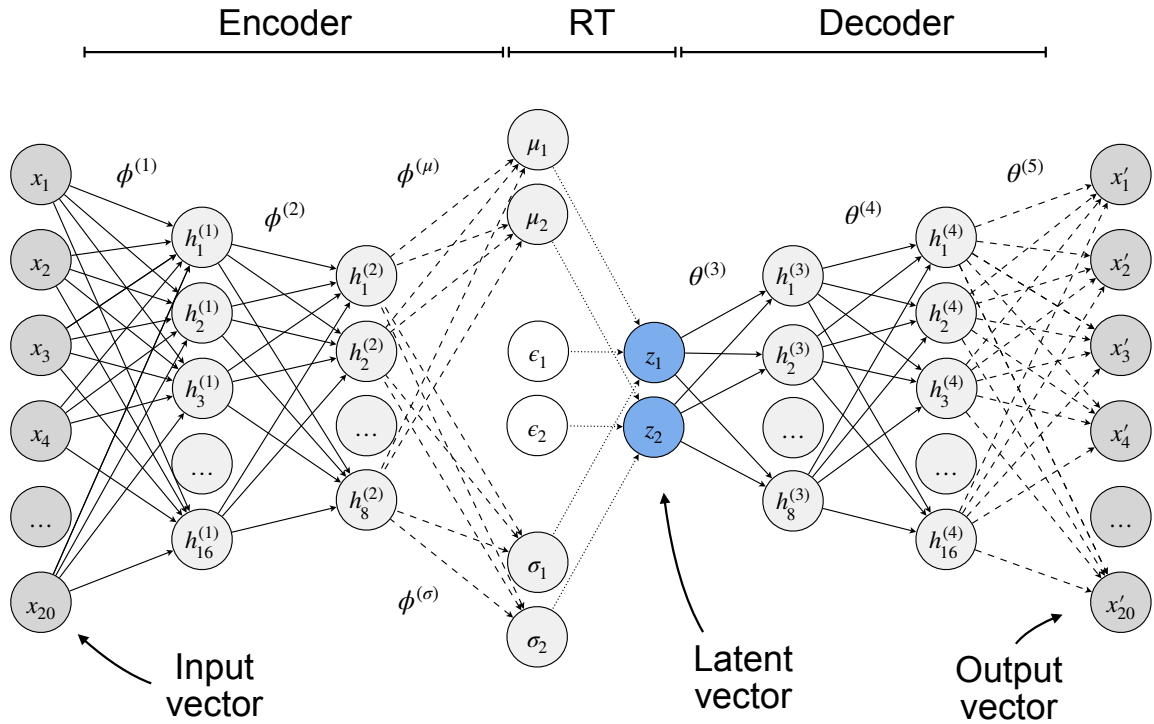
{Network architecture of the variational autoencoder used to map induced polarization data input vectors ($\mathbf{x}$) to latent vector samples ($\mathbf{z}$) and back to a reconstruction of the input data ($\mathbf{x'}$). All hidden layers ($\mathbf{h}^{(l)}$) are fully connected by linear transformation parameters $\phi^{(l)}$ in the encoder and $\theta^{(l)}$ in the decoder. Solid arrows indicate the use of a hyperbolic tangent activation function after the linear transformation. Dashed arrows indicate a linear transformation with no activation function. Dotted arrows indicate the reparametrization trick (RT).
}

We trained the IP-VAE on $\mathcal{D}_\textrm{Train}$, which contained 1\,600\,319 examples of IP decays. Each IP decay served as both the input and output of the network to achieve fully unsupervised learning. The trainable network parameters $\phi$ and $\theta$ were optimized using the Adam method \citep{kingma_adam_2014} and an initial learning rate of $10^{-3}$. 47~726 optimization steps were required to complete one training epoch using a mini-batch size of 32. The completion of one epoch means that every sample in $\mathcal{D}_\textrm{Train}$ served to update the trainable parameters of the neural network \citep{goodfellow_deep_2016}. Figure~\ref{fig:learning-curves-zdim=2} shows the evolution of $\mathcal{L}_{\textrm{VAE}}$ and its individual $D_{\textrm{KL}}$ and NLL terms. The optimizer reached a stationary state after one training epoch, and the completion of additional epochs did not improve $\mathcal{L}_{\textrm{VAE}}$. It is not uncommon for neural networks to reach an optimal state after only one epoch when performing unsupervised learning tasks on extensive data sets \citep{komatsuzaki_one_2019}.

\plot{learning-curves-zdim=2}{width=0.8\textwidth}
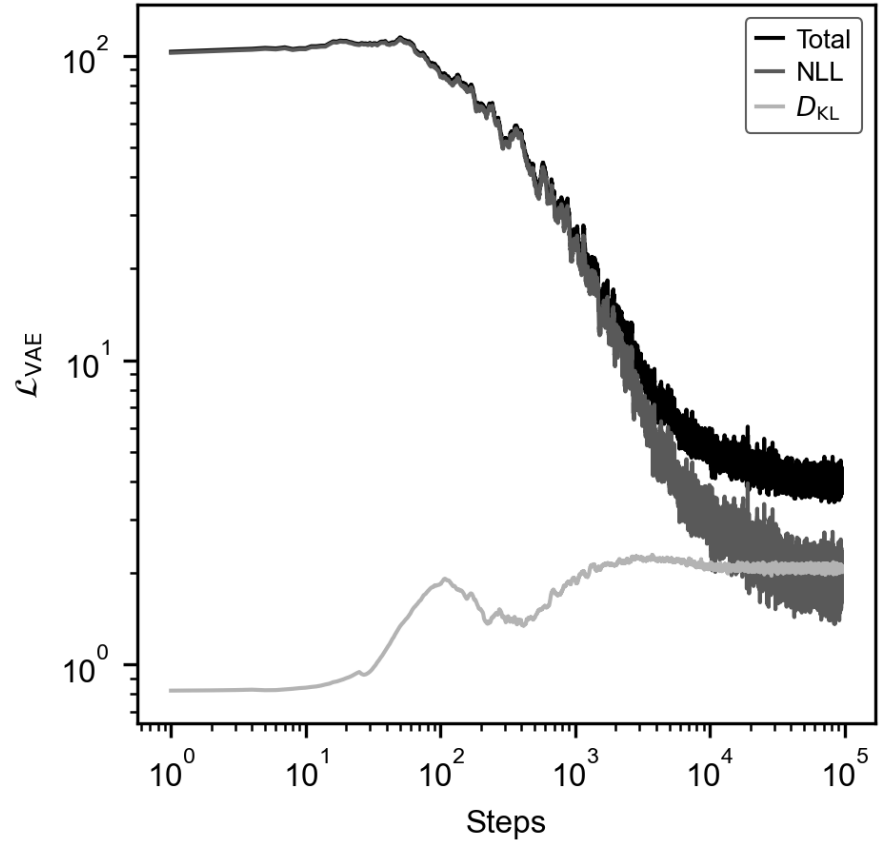
{Learning curves of the IP-VAE for the first two training epochs. After 47~726 optimization steps, corresponding to the end of the first training epoch, the negative log-likelihood (NLL) and the Kullback–Leibler divergence ($D_{\textrm{KL}}$) reached stationary values. The $D_{\textrm{KL}}$ and NLL were well-balanced and converged to similar values at the end of the optimization process. The total variational autoencoder loss ($\mathcal{L}_{\textrm{VAE}}$) is the sum of the $D_{\textrm{KL}}$ and NLL terms.
}

\subsection{Baseline denoising methods}
We used three conceptually simple filtering methods to establish denoising baselines and evaluate the performance of our IP-VAE approach: a moving average, an exponential moving average, and a Butterworth filter. However, these methods required tuning one or several hyperparameters to achieve the best denoising effect possible. Furthermore, we needed to optimize the hyperparameters based on each decay curve, which significantly reduced the practicality of these methods when processing $\mathcal{D}_{\textrm{STest}}$ and $\mathcal{D}_{\textrm{FTest}}$. 

\subsubsection{Simple moving average}
The moving average (MA) is a finite impulse response filter and perhaps one of the simplest approaches to denoising sequential data. The mathematical formulation of a $M$-order MA filter is
\begin{equation}
    x_j' = \frac{1}{M} \sum_{s=-S}^S x_{j+s},
\end{equation}
where $x_j$ refers to the $j$\textsuperscript{th} element of the input sequence, $x_j'$ is its filtered version and $M=2S+1$. We padded the IP readings to preserve values at the boundaries when applying the MA filter.

\subsubsection{Exponential moving average}
The exponential moving average (EMA) is a first-order infinite impulse response filter where the weighting of each data point decreases exponentially. The EMA depends on a real hyperparameter $\alpha \in [0, 1]$, which must be tuned to obtain satisfying denoising results. The EMA filter is computed recursively and defined as
\begin{equation}
  x'_j=\begin{cases}
    x_1, & \text{if $j=1$}.\\
    \alpha x_j + (1 - \alpha)x'_{j-1}, & \text{otherwise}.
  \end{cases}
\end{equation}
Edge cases of the EMA are $\alpha=1$ (no denoising) and $\alpha=0$, for which the result is a constant equal to the first value of the sequence.

\subsubsection{Butterworth filter}
Butterworth filters have a maximally flat frequency response in their passband. We used a first-order low-pass Butterworth filter, for which the frequency response is given by 
\begin{equation}
    G(\omega_\textrm{n}) = \frac{1}{\sqrt{1 + \omega_\textrm{n}^2}},
\end{equation}
where $\omega_\textrm{n} \in [0, 1]$ is the cutoff frequency normalized by the Nyquist frequency.

\subsection{Reconstruction quality metrics}
We used two metrics to evaluate the reconstruction and denoising performance of the IP-VAE and other baseline denoising methods: the root-mean-square error (RMSE) and the peak signal-to-noise ratio (S/N).

\subsubsection{Root mean square error}
The RMSE corresponds to the Euclidean norm of the misfit between the input data and the reconstructed data. For IP data, RMSE is expressed in mV/V and is defined as
\begin{equation}
    \begin{split}
        \textrm{RMSE} &= \left\lVert \mathbf{x}-\mathbf{x'} \right\rVert_2  \\
        &= \left(\frac{1}{d}\sum_j^d \lv x_j - x_j' \rv^2 \right)^\frac{1}{2},
    \end{split} 
\end{equation}
where $d$ corresponds to the number of chargeability integration windows per measurement cycle ($d = 20$). The RMSE is scale-dependent, which can make its interpretation more difficult when comparing various IP readings. 

\subsubsection{Peak signal-to-noise ratio}
The peak S/N computation involves the dynamic range of the data and this metric is less dependent on the data scale than the RMSE. Peak S/N is expressed in decibels (dB) and is defined as 
\begin{equation}
    \textrm{Peak S/N} = 20 \log_{10}{\left( \frac{\max{\mathbf{x}} - \min{\mathbf{x}}}{\left\lVert \mathbf{x}-\mathbf{x'} \right\rVert_2} \right)}.
\label{eq:psnr}
\end{equation}

\section{Results}

In this section, we present the various applications of the IP-VAE. First, we demonstrate the generative properties of the probabilistic decoder by creating a synthetic data set which is a close match to the actual data compiled in $\mathcal{D}_\textrm{Train}$. Second, we demonstrate the denoising application of the IP-VAE in two parts: (a) by contaminating $\mathcal{D}_\textrm{STest}$ with additive Gaussian white noise and denoising it, and (b) by denoising the field data comprised in $\mathcal{D}_\textrm{FTest}$. We present both qualitative and quantitative evaluations of the denoising capabilities of the IP-VAE for the synthetic and natural data sets and compare the results with those of the baseline filtering methods. Third, we demonstrate how the IP-VAE reconstruction peak S/N is interpretable in terms of overall survey quality. Last, we show that the IP-VAE reconstruction RMSE is applicable as an outlier score consistent with an empirical confidence score determined independently at the time of survey completion.

\subsection{Data generation}

The first application of the IP-VAE is synthetic data generation. The generative properties of VAE are well-documented, but care is needed to ensure that the learned latent representation allows reconstruction of the entire data space, not just the data average. We compared the synthetic data contained in $\mathcal{D}_{\textrm{STest}}$ with the natural data in $\mathcal{D}_{\textrm{Train}}$ using density line charts \citep{moritz_visualizing_2018}. Figure~\ref{fig:data-distribution-zdim=2} shows the comparison between the real input data distribution and the synthetic data distribution decoded from $\mathbf{z} \sim \mathcal{N}(0, 1)$. 

\plot{data-distribution-zdim=2}{width=\textwidth}
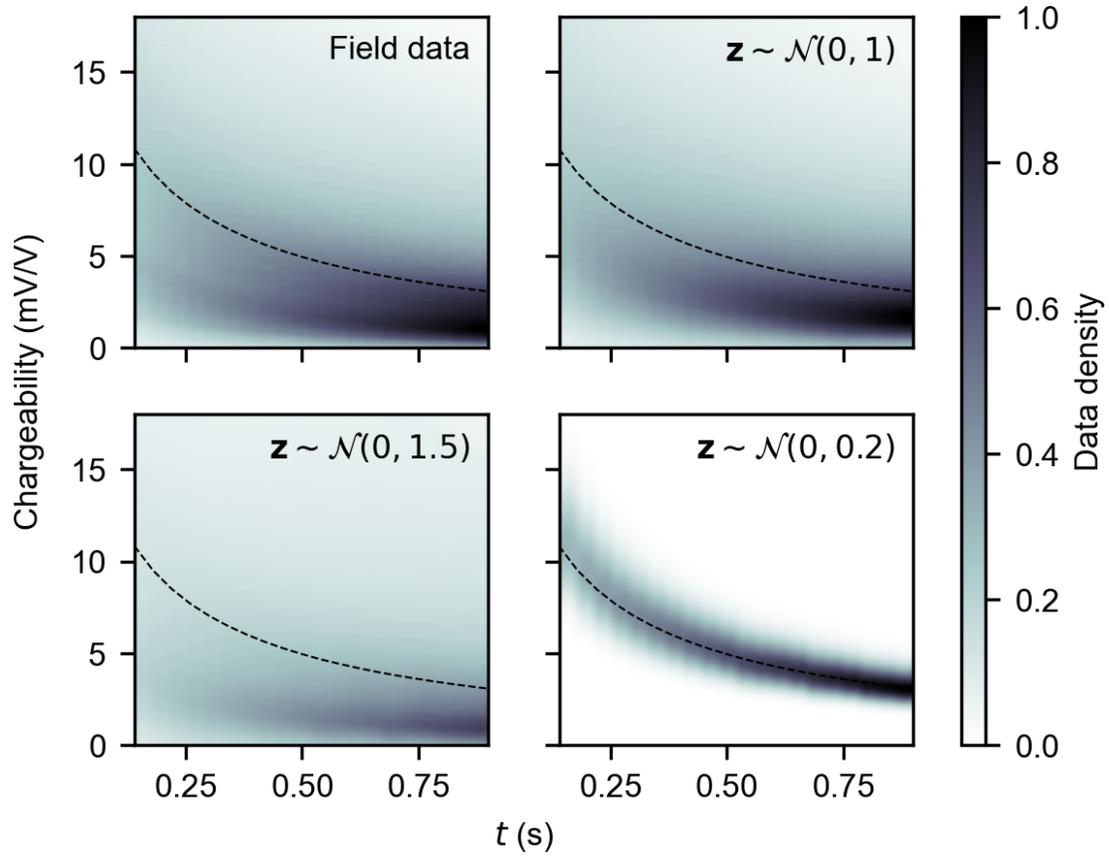
{Comparison between the field data space (input) and the synthetic data space (output) generated by decoding 1.6M latent vector samples ($\mathbf{z}$). Decoding samples drawn from the standard normal distribution $\mathcal{N}(0, 1)$ yielded an excellent representation of the data space. In comparison, decoding samples from $\mathcal{N}(0, 1.5)$ generated a flared data space, and decoding samples from $\mathcal{N}(0, 0.2)$ only produced decays that closely resembled the IP data compilation median (drawn as the dashed line). 
}

The similarities between the real and synthetic data distributions are striking. We concluded that the latent representation learned by the IP-VAE allowed the decoder to express IP data curves that are closely similar to those of the natural data space. Data obtained by decoding $\mathbf{z} \sim \mathcal{N}(0, 1.5)$ and $\mathbf{z} \sim \mathcal{N}(0, 0.2)$ are also shown in Figure~\ref{fig:data-distribution-zdim=2} to illustrate how sampling from wider and narrower distributions generated flared and specific data sets, respectively. Synthetic data converged toward the median of the real data as $\sigma\to 0$. Decoding large values of $\mathbf{z}$ (e.g., $\gg 1$) naturally produced inconsistent results because the $D_\textrm{KL}$ regularization term favors $\mathbf{z}\sim \mathcal{N}(0, 1)$ during the training phase. 

\subsection{Bayesian denoising}

The second application of the IP-VAE is Bayesian denoising. The VAE latent space acts as an information bottleneck. As a result, the outputs of the IP-VAE were effectively smoother versions of their corresponding inputs. Moreover, the stochastic nature of the VAE allows for multiple reconstruction realizations of each input IP sequence, thereby estimating the reconstruction distribution. The 0.5, 0.025, and 0.975 quantiles were computed from this distribution and used to describe the best fit and its 95\% confidence interval. This section reports the results of our Bayesian denoising experiments conducted on synthetic and natural data. Thus, we provide both quantitative and qualitative indicators of the performance of the IP-VAE as an unsupervised denoising technique for IP data. Note that IP-VAE forward passes are computationally inexpensive (see Table~\ref{tbl:dtrain}).  

\subsubsection{Synthetic data}

$\mathcal{D}_{\textrm{STest}}$ consists of 1.6M synthetic IP decays that were generated by decoding $\mathbf{z} \sim \mathcal{N}(0, 1)$ with the IP-VAE. Synthetic data generated using this method were shown to be representative of the real data used for model training (Figure~\ref{fig:data-distribution-zdim=2}) and are considered ground truth values in the following tests. Additive Gaussian white noise with varying standard deviations ($\sigma_\textrm{noise}$, in mV/V) contaminated the ground truth synthetic data. For the qualitative test, the IP-VAE processed the noisy synthetic data examples, and we inspected their reconstructed outputs visually. Figure~\ref{fig:2496} shows an example of a noisy synthetic IP decay before and after denoising by the IP-VAE and the three baseline methods. Qualitatively, the IP-VAE provided the best denoising outputs compared to the baseline methods for all synthetic data that we inspected.

\plot{2496}{width=\textwidth}
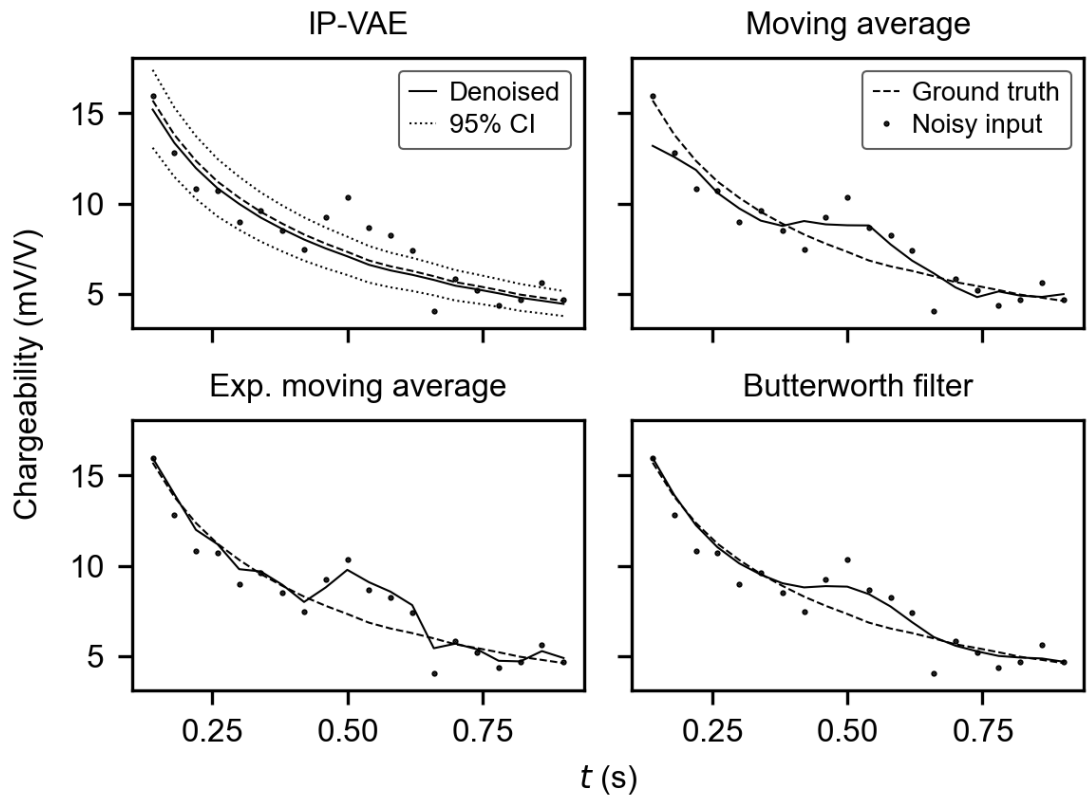
{Example of denoising results for a synthetic ground truth IP decay generated by the decoder. The dot markers represent the ground truth IP data after contamination with additive Gaussian white noise of standard deviation 1.1 mV/V. The results of four denoising techniques are compared: the IP-VAE, a moving average, an exponential moving average and a first-order Butterworth filter. CI: confidence interval.
}

The baseline methods also provided denoising capabilities to some extent, but their denoised outputs yielded curves that are not necessarily in agreement with the typical shape of an IP decay curve. Moreover, the baselines methods could not remove data spikes as they were more sensitive to outlying chargeability windows, as evidenced by the denoised results around 0.5 s in Figure~\ref{fig:2496}. In summary, the IP-VAE was the only method with spike removal capabilities, and that provided denoised outputs closely similar to the ground truth synthetic data. It is also noteworthy that the IP-VAE approach provided an estimate of the data uncertainty through 100 denoising realizations of the same input data. As a result, the IP-VAE 95\% confidence interval is highlighted with the dotted lines in Figure~\ref{fig:2496} and is a reasonable estimate compared to the spread of the noise-contaminated data. 

We computed the RMSE between the 1.6M ground truth synthetic data and their denoised versions for the quantitative test. Table~\ref{tbl:sftest} summarizes the denoising performance of the IP-VAE and baseline methods for the entire synthetic test data set with $\sigma_\textrm{noise} = 1.1$ mV/V.

\tabl{sftest}{Comparison between baseline denoising methods and the variational autoencoder (IP-VAE). MA is a moving average. EMA is an exponential moving average. $\mathcal{D}_\textrm{STest}^\textrm{RMSE}$ is the denoising root-mean-square error for the synthetic test data set (contaminated by additive Gaussian white noise with a standard deviation of 1.1 mV/V). $\mathcal{D}_\textrm{FTest}^\textrm{RMSE}$ is the denoising root-mean-square error for the field test data set. The given nominal values and uncertainties respectively correspond to the means and standard deviations over the entire data sets. Boldface highlights the optimal results. 
}{
\begin{center}
    \begin{tabular}{ccc}
        \hline
        Method & $\mathcal{D}_\textrm{STest}^\textrm{RMSE}$ & $\mathcal{D}_\textrm{FTest}^\textrm{RMSE}$ \\
        \hline
        \hline
        None & $4.859 \pm 0.771$ & $6.138 \pm 6.820$ \\
        \hline 
        IP-VAE & $\mathbf{1.352 \pm 1.030}$ & $\mathbf{2.917 \pm 3.140}$ \\
        \hline
        MA & $2.964 \pm 1.104$ & $4.386 \pm 4.250$ \\
        \hline
        EMA & $3.889 \pm 0.988$ & $5.061 \pm 5.047$ \\
        \hline
        Butterworth & $2.627 \pm 0.852$ & $4.249 \pm 4.846$ \\
        \hline
    \end{tabular}
\end{center}
}

Results in Table~\ref{tbl:sftest} are further evidence that the IP-VAE outperformed the baseline methods. In addition, the IP-VAE denoising capabilities were more robust to varying noise levels than the baseline methods. As summarized in Figure~\ref{fig:noise-sensitivity}, increasing values of $\sigma_\textrm{noise}$ were proportional to the RMSE for all denoising methods. However, for all $\sigma_\textrm{noise}$ values between 0 and 3 mV/V, the IP-VAE outperformed the three baseline methods at the denoising task. The IP-VAE also adopts the line with the minimal slope in Figure~\ref{fig:noise-sensitivity}, indicating that it is more robust to high noise levels than the baseline methods. Interestingly, the exponential moving average filter had the worst denoising performance of all methods. Also noteworthy is the performance of the simple moving average filter, which was greater than that of the more sophisticated Butterworth filter for $\sigma_\textrm{noise} > 2$ mV/V. 

\plot{noise-sensitivity}{width=\textwidth}
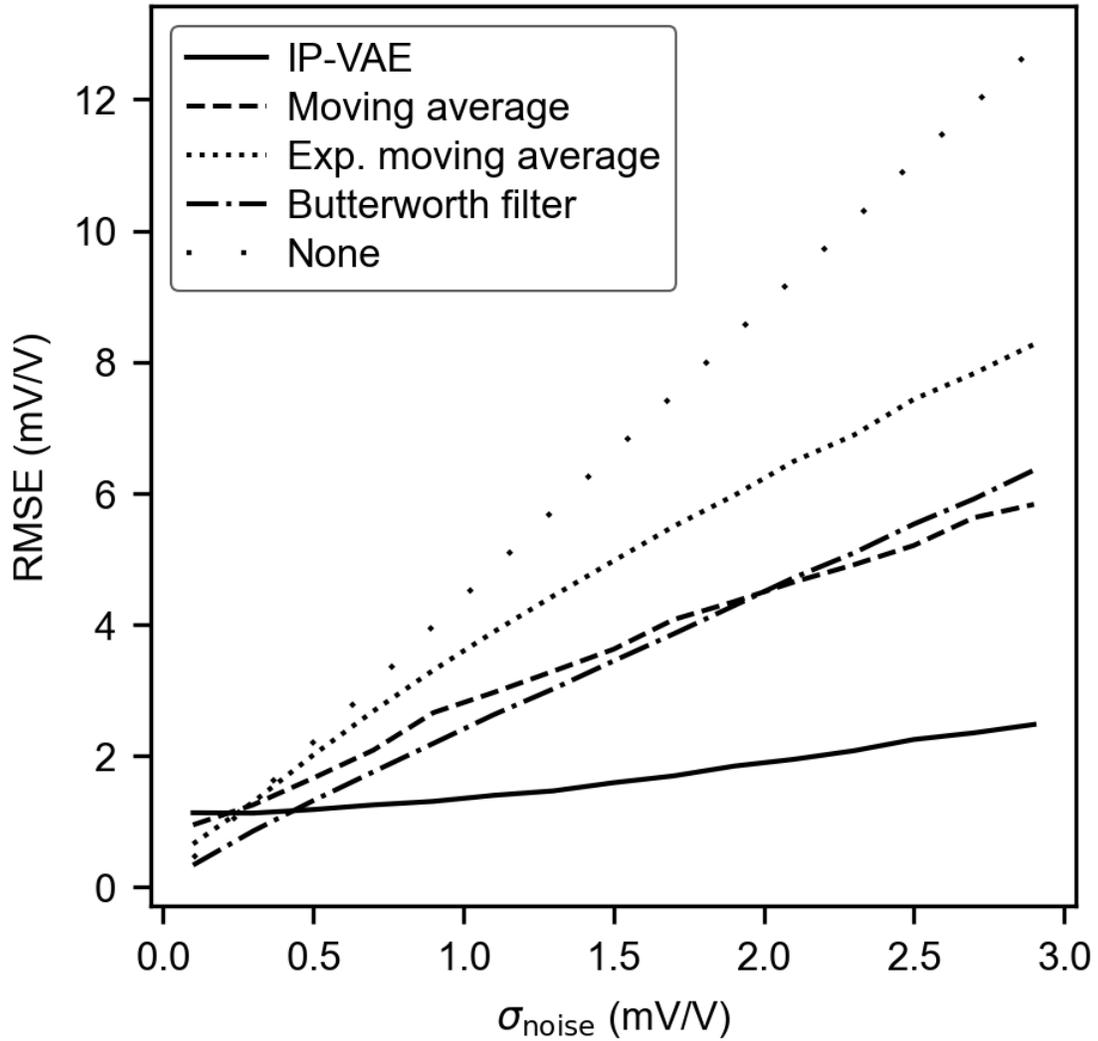
{Reconstruction root-mean-square error (RMSE) as a function of the synthetic Gaussian noise standard deviation ($\sigma_\textrm{noise}$) for various denoising methods. The IP-VAE significantly outperformed all other methods for $\sigma_\textrm{noise}$ values above 0.4 mV/V.
}

\subsubsection{Field data}

For our first qualitative test on field data, we inspected the IP-VAE outputs and compared them with their respective inputs from $\mathcal{D}_{\textrm{Train}}$ after the unsupervised training phase. Figure~\ref{fig:field-examples-zdim=2-781974-484961-572179} shows three examples of IP field measurements from $\mathcal{D}_{\textrm{Train}}$ after training the IP-VAE. These examples represent varying peak S/N and reconstruction errors.

\plot{field-examples-zdim=2-781974-484961-572179}{width=0.9\textwidth}
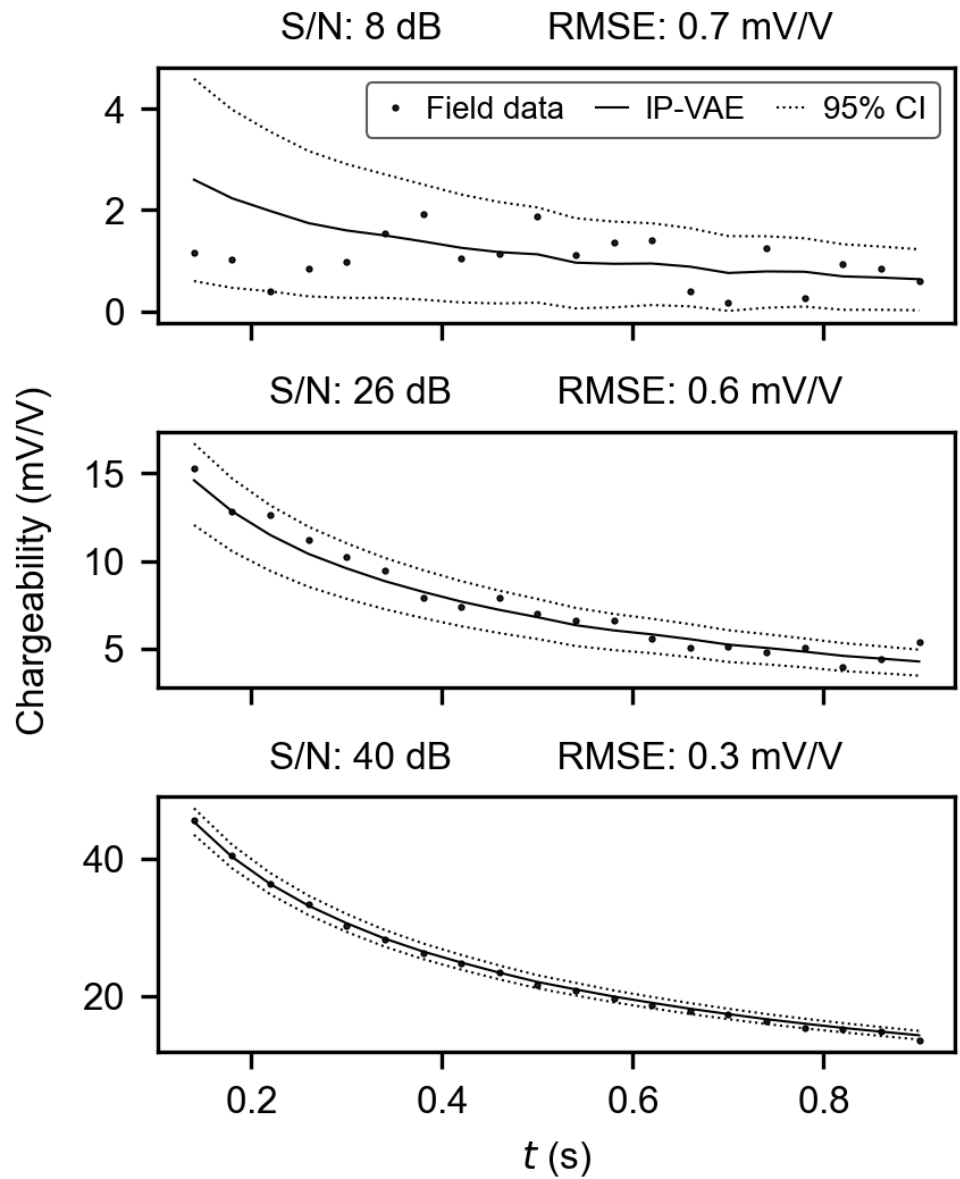
{Representative examples from the training data set with their respective IP-VAE outputs. Examples are presented in ascending order of peak signal-to-noise ratio (S/N) from top to bottom. RMSE: root-mean-square error. CI: confidence interval.
}

The reconstructed examples shown in Figure~\ref{fig:field-examples-zdim=2-781974-484961-572179} are a good fit with the input field data, and the inferred IP-VAE 95\% confidence interval was representative of the noisy input data spread. Furthermore, the peak S/N and RMSE values given in Figure~\ref{fig:field-examples-zdim=2-781974-484961-572179} were a good representation of the overall data quality. For example, the IP curve with 40 dB peak S/N and 0.3 mV/V RMSE is a representative sample of some of the highest quality measurements in $\mathcal{D}_{\textrm{Train}}$. On the other hand, the IP curve characterized by 8 dB peak S/N and 0.7 mV/V RMSE is a representative sample from the lowest S/N IP decays in $\mathcal{D}_{\textrm{Train}}$. The IP-VAE was robust to outliers and provided smooth IP decay reconstructions of all training examples. 

For the following quantitative test, we considered the high S/N versions of IP decays in $\mathcal{D}_{\textrm{FTest}}$ as ground truth reference curves. For example, figure~\ref {fig:comparison-field-denoise-28} shows a denoising result from $\mathcal{D}_{\textrm{FTest}}$ before and after processing the low S/N versions of the measurements with the IP-VAE and baseline methods.

\plot{comparison-field-denoise-28}{width=\textwidth}
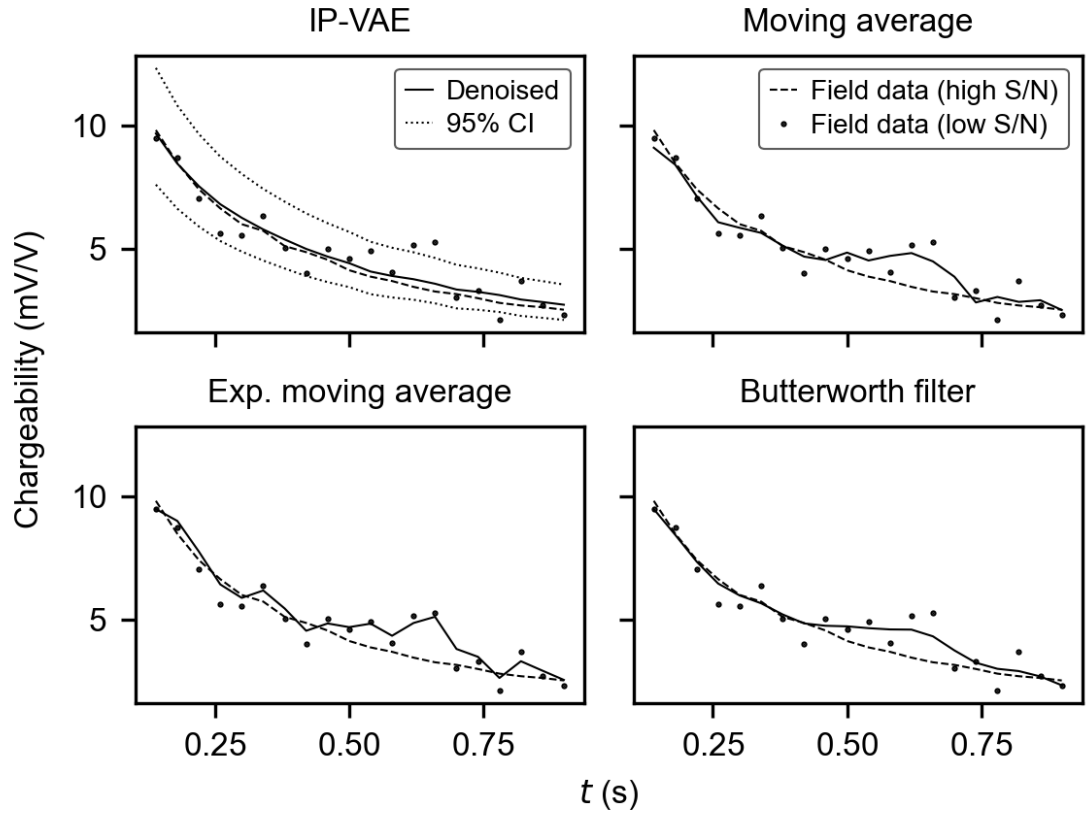
{Comparison of denoising results from the field test data set using the IP-VAE, a moving average, an exponential moving average, and a Butterworth filter. Displayed are the high and low signal-to-noise ratio (S/N) versions of the same decay, obtained respectively with injected currents of 2.09 A and 0.22 A. Each denoising method aimed to process the low S/N version to recover the high S/N version. CI: confidence interval.
}

Consistently with our test results on synthetic data, the IP-VAE qualitatively outperformed the baseline methods at the denoising task on all field measurements that we inspected. The IP-VAE outputs obtained by denoising the low S/N (low current) data closely matched their high S/N (high current) counterparts. Furthermore, the baseline methods were more sensitive to outliers than the IP-VAE when attempting to denoise the signal, as visible near the spike in the 0.5 to 0.75 s range of Figure~\ref{fig:comparison-field-denoise-28}. Table~\ref{tbl:sftest} provides a quantitative summary of the IP-VAE and baseline methods denoising performances for $\mathcal{D}_{\textrm{FTest}}$.

\subsubsection{Adjustable denoising}
The $D_\textrm{KL}$ regularization term in $\mathcal{L}_\textrm{VAE}$ and the data compression in the IP-VAE latent space are responsible for smoothing the IP-VAE reconstructions. There are two ways to tune the denoising strength of the IP-VAE. The first is by modifying the latent space dimensions. High-dimensional latent spaces can capture all the noise components in data, whereas low-dimensional latent spaces can only encode the general shape of the data (i.e., the pure IP decay). The second is by weighting $D_\textrm{KL}$ relative to the reconstruction error in the computation of $\mathcal{L}_\textrm{VAE}$ \citep[beta-VAE,][]{higgins_beta-vae_2017}. Increasing the weight of the $D_\textrm{KL}$ term yielded smoother reconstructions. However, excessive weight on the $D_\textrm{KL}$ term made all reconstructions resemble the mean of $\mathcal{D}_\textrm{Train}$. We obtained all the denoising results using a two-dimensional latent space and without weighting the $D_\textrm{KL}$ term.

\subsection{Survey S/N analysis}

The third application of the IP-VAE is related to data quality control, specifically survey S/N analysis and interpretation. The peak S/N reconstruction error is a good indicator of the overall measurement quality because noise-free IP curves will be reconstructed almost perfectly by the IP-VAE. On the other hand, the reconstruction error will naturally be higher when the input data is noisy due to the denoising properties of the IP-VAE. Evaluation of the noise level is possible on a survey-by-survey basis or entire collections of IP data. Figure~\ref{fig:psnr-zdim=2} demonstrates the use of the IP-VAE as a tool to compare the overall quality of different surveys, which is interpretable based on field conditions impacting $V_\textrm{p}$ and contact resistance between the potential electrodes.

\plot{psnr-zdim=2}{width=0.9\textwidth}
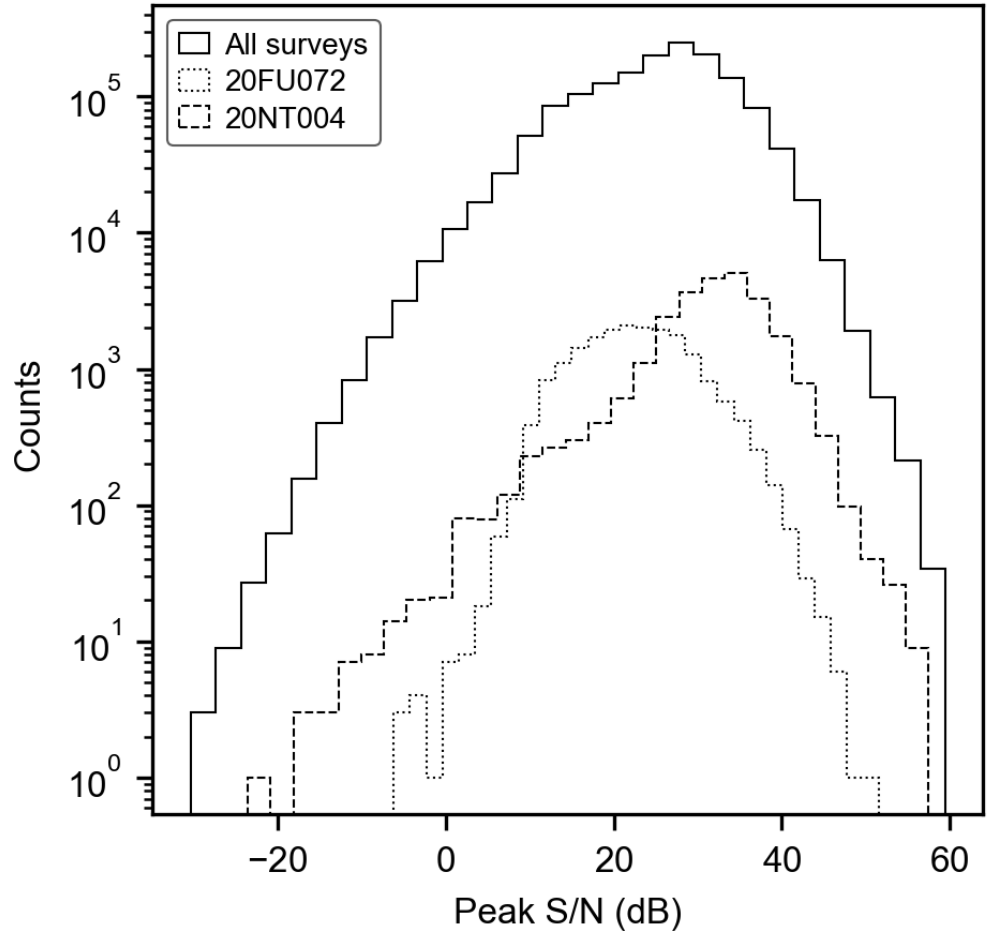
{Histogram of peak signal-to-noise ratio (S/N) evaluated by the IP-VAE for all measurements included in the training IP data compilation ($\mathcal{D}_\textrm{Train}$). Also displayed are peak S/N histograms of two individual surveys from different geolocations. The peak S/N of survey 20NT004 (Wawa Gold Camp, Ontario) was approximately twice that of survey 20FU072 (Kazakhstan). The IP-VAE peak S/N estimations were consistent with other quality metrics obtained independently.
}

In Figure~\ref{fig:psnr-zdim=2}, the peak S/N histograms of surveys 20NT004 (Wawa Gold Camp, Ontario) and 20FU072 (Kazakhstan) are plotted and compared with the histogram of the entire survey compilation. One interpretation is that survey 20NT004 had vastly superior S/N than survey 20FU072, as evidenced by their S/N histogram peaks at approximately 40 and 20 dB, respectively. This interpretation is consistent with the metrics given in Table~\ref{tbl:dtrain}. The subsurface investigated in 20FU072 was 40 times more conductive than that of 20NT004. Consequently, the average measured $V_\textrm{p}$ was significantly lower (36.73 mV) for 20FU072 than it was for 20NT004 (633 mV), despite similar values of injected current (2079 and 890 mA, respectively). Table~\ref{tbl:dtrain} also reveals that survey 20NT004 had an overall higher empirical confidence score (defined in the next section). However, whereas 20NT004 had an overall higher S/N than 20FU072, it also contained more measurements with unacceptable negative values of S/N. An interpretation for the overrepresentation of negative S/N values in 20NT004 is that establishing a low contact resistance at the potential electrode was more challenging to achieve in the field conditions of the Wawa Gold Camp than it was for the survey conducted in Kazakhstan (see Table~\ref{tbl:dtrain}). 

\subsection{Outlier detection}
The fourth application of the IP-VAE is outlier detection. The RMSE reconstruction metric is often used as an outlier score when processing data with VAE. Outlier data are, by definition, rare and conspicuous deviations from the data population. Consequently, outlying measurements did not hold much weight when the IP-VAE learned to minimize the loss function across $\mathcal{D}_\textrm{Train}$. The IP-VAE thus produced high reconstruction RMSE when processing outlying decays. To evaluate the application of RMSE as an outlier score, we compared it with an empirical confidence score attributed to each measurement during a previous round of semi-automated data inspection shortly after the completion of each survey. The confidence score is expressed in percentage and depends on the primary voltage, electrode contact resistance, and quantitative comparison of the decay curve with the normalized decay programmed in the ELREC Pro receiver, among other indicators. The empirical confidence score is entirely independent of the IP-VAE reconstruction metrics. Figure~\ref{fig:anomalies} shows a two-dimensional scatter plot of the peak S/N and RMSE reconstruction metrics with a color-coded empirical confidence score.

\plot{anomalies}{width=\textwidth}
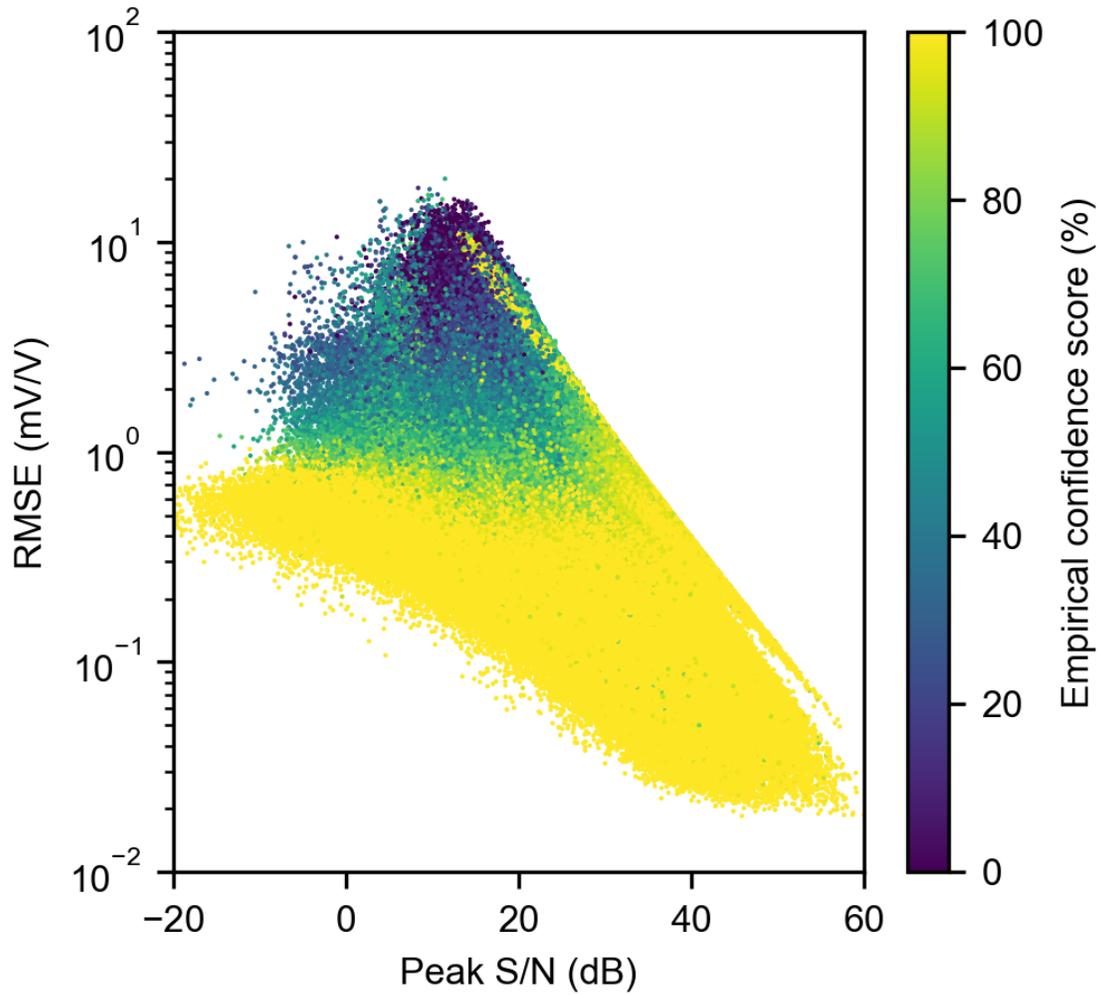
{Scatter plot showing the relationship between the reconstruction root mean square error (RMSE), the peak signal-to-noise ratio (S/N), and a confidence score attributed empirically to each measurement during semi-automated processing. A 1 mV/V RMSE threshold could automatically flag IP decays with a low confidence score.
}

There is good agreement between the reconstruction RMSE and the empirical confidence score as illustrated in Figure~\ref{fig:anomalies}. In addition, we observed that the IP measurements initially flagged with a confidence score below 50\% were characterized by RMSE values above one mV/V. Thus, our results suggest that simple thresholding of the IP-VAE reconstruction RMSE is an excellent strategy to flag low-confidence IP measurements automatically for further inspection. However, we also observed that several IP measurements suffered from high reconstruction RMSE and low peak S/N despite being labeled with a confidence score of 100\%. Mislabeling due to human error may explain this discrepancy, and closer inspection to re-assess the confidence score of these IP readings is needed.

\section{Interpretation of the latent space}

\subsection{Correlation with average chargeability}
The IP-VAE latent space has no intrinsic physical meaning. Therefore it is of interest to interpret it in terms of more specific parameters that do carry a physical meaning (e.g., $\bar{m}$). Figure~\ref{fig:2d-latent-space-zdim=2} shows a scatter plot of the two-dimensional latent space representing data from the entire survey compilation with color-coded $\bar{m}$. 

\plot{2d-latent-space-zdim=2}{width=\textwidth}
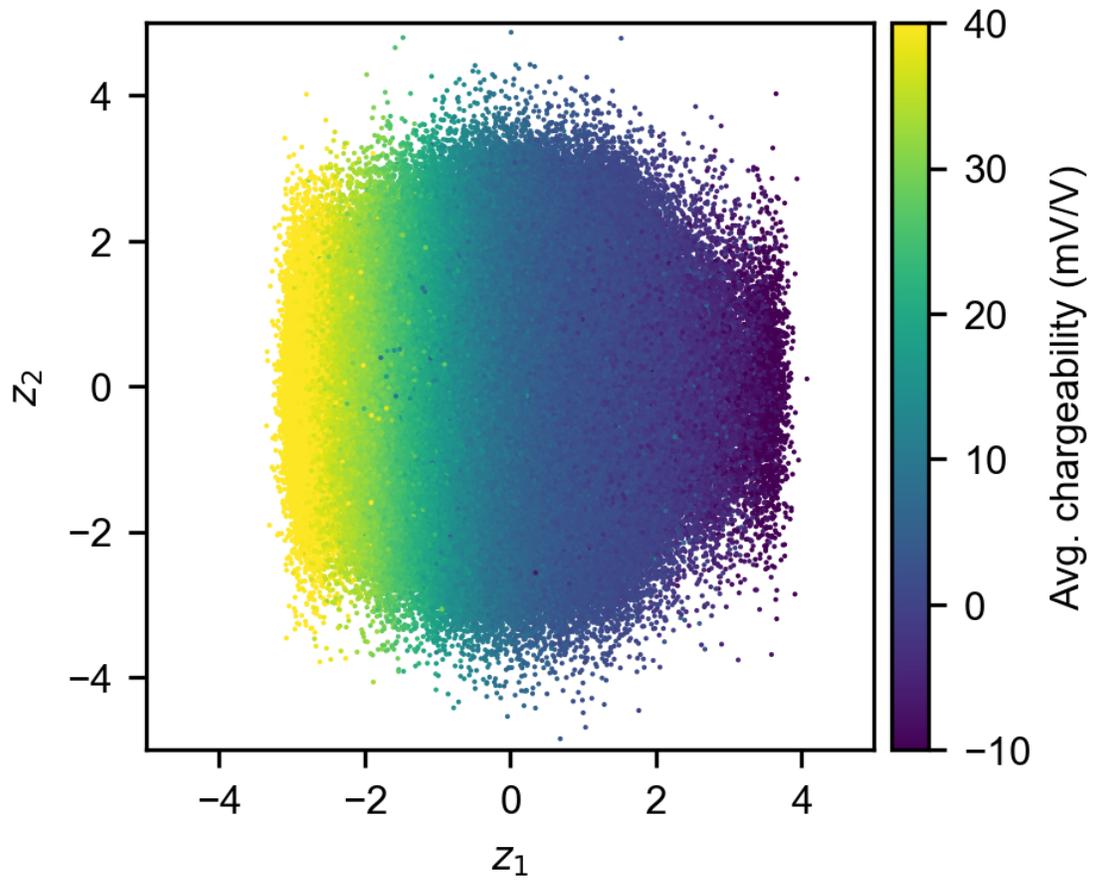
{Two-dimensional scatter plot representation of the IP-VAE latent space. The scatter plot represents an embedding of 1\,600\,319 IP curves. The first dimension of the latent vectors ($z_1$) is an excellent representation of the average chargeability. 
}

We observed that the average chargeability parameter carries the meaning of the latent vectors' first dimension ($z_1$). However, we did not find a simple interpretation for the second dimension ($z_2$) by comparing it with other IP parameters, including the dynamic range, average decay slope, and IP decay mid-point slope. This result hints that the second dimension of the latent vectors is not necessary to reconstruct noise-free IP observations. Instead, it could serve to capture more subtle information about the noise in the data. We further explore this hypothesis in the following subsection.

\subsection{Influence of latent space dimensions}
Ideal parametrization of the IP phenomenon is still an open question. Specifically, there is no known bound on the minimal number of parameters required to fully describe IP data using a mathematical model and reconstruct it with minimal loss of information. We evaluated the average reconstruction RSME and peak S/N metrics of four models with latent space dimensions $k \in \{1, 2, 4, 6\}$ to establish how many parameters were needed to encode IP data. We also generated four synthetic data sets by decoding $\mathbf{z} \sim \mathcal{N}(0, \mathbf{I}_k)$, where $k \in \{1, 2, 4, 6\}$, to compute the absolute difference between their respective density line charts \citep{moritz_visualizing_2018} and that of $\mathcal{D}_{\textrm{Train}}$. Table~\ref{tbl:latent-comparison} summarizes the results.

\tabl{latent-comparison}{Comparison of the training losses, reconstruction metrics, test scores, and generative properties of the IP-VAE with varying latent space dimensions ($K$). NLL: negative log-likelihood. $D_\textrm{KL}$: Kullback-Leibler divergence. $\mathcal{D}_\textrm{Train}^\textrm{S/N}$: mean reconstruction peak signal-to-noise ratio of the training set. $\mathcal{D}_\textrm{Train}^\textrm{RMSE}$: mean reconstruction root-mean-square error of the training set. $\mathcal{D}_\textrm{FTest}^\textrm{RMSE}$: mean reconstruction root-mean-square error of the field test set. $\lv\mathcal{D}_\textrm{STest}^\textrm{DLC}-\mathcal{D}_\textrm{Train}^\textrm{DLC}\rv$: absolute difference between the density line charts of the training set and synthetic test sets. Boldface highlights optimal results.
}{
\begin{center}
    \begin{tabular}{ccccccc}
        \hline
        $K$ & NLL & $D_\textrm{KL}$ & $\mathcal{D}_\textrm{Train}^\textrm{S/N}$ (dB) & $\mathcal{D}_\textrm{Train}^\textrm{RMSE}$ (mV/V) & $\mathcal{D}_\textrm{FTest}^\textrm{RMSE}$ (mV/V) & $\lv\mathcal{D}_\textrm{STest}^\textrm{DLC}-\mathcal{D}_\textrm{Train}^\textrm{DLC}\rv$  \\
        \hline
        \hline
        1 & \textbf{1.913} & 2.071 & 24.290 & 0.7853 & 3.459 & \textbf{0.0037} \\
        \hline
        2 & 1.916 & 2.071 & \textbf{25.053} & \textbf{0.6939} & \textbf{3.124} & 0.0045 \\
        \hline
        4 & 1.924 & 2.071 & 24.581 & 0.7524 & 3.249 & 0.0042 \\
        \hline
        6 & 1.916 & 2.072 & 23.698 & 0.7943 & 3.246 & 0.0042 \\
        \hline
    \end{tabular}
\end{center}
}

As evidenced by the results in Table~\ref{tbl:latent-comparison}, there was only a marginal difference in the denoising and generative properties of the IP-VAE when using varying latent dimensions. Using $k=1$ yielded the lowest training losses and best generative capabilities. Using $k=2$ yielded the best reconstruction metrics over $\mathcal{D}_\textrm{Train}$ and the best denoising capabilities for $\mathcal{D}_\textrm{FTest}$, thus justifying our choice of $k=2$ for the final IP-VAE. We reiterate that the performance difference between the models was marginal and that all four models had excellent generative modeling properties and denoising performances.

\section{Discussion}

\subsection{Related work}
We demonstrated that the IP-VAE outperforms baseline denoising methods. However, more advanced and effective time-domain IP signal processing methods already exist \citep{olsson_doubling_2016, flores_orozco_decay_2018, barfod_automatic_2021}. Comparing the performance of these methods and the IP-VAE is outside the scope of this paper because denoising is not the sole motivation for developing each method. Instead, we compare other properties of the IP-VAE with those of previously published methods and highlight some aspects which make it an interesting tool for IP researchers and practitioners. It is worth noting that implementations of the three aforementioned methods are not openly available, whereas we released a pre-trained IP-VAE as an open-source Python package. Also, the generative IP modeling properties are a unique feature of the IP-VAE.

\cite{olsson_doubling_2016} introduced a strategy to remove background drift from telluric or instrumental sources, voltage spikes from anthropogenic or natural sources (e.g., electric fences and lightning), harmonic noise caused by the instrument power supplies, and Gaussian white noise. While this strategy is comprehensive and significantly extends the frequency content of IP data in preparation for Cole-Cole inversion, it is also rigid and relies on simplifying assumptions regarding what constitutes a standard IP reading. Expressly, the approach to drift removal therein assumes a Cole-Cole model relaxation, and the data uncertainty estimation protocol relies on computing the standard deviation of misfit between a convolution of the data and an exponential function. In comparison, the proposed IP-VAE is entirely data-driven and agnostic to Cole-Cole or exponential models. We do not assume the standard shape of IP decays but instead learn to approximate it from an extensive collection of field observations. Moreover, the IP-VAE approach allows estimating data uncertainty within a proper Bayesian framework rather than a deterministic one.

\cite{flores_orozco_decay_2018} proposed a rule-based decay curve analysis method. An important motivation for the contribution therein is to accelerate the IP data collection process by removing the need for reciprocal electrode measurements. The authors also introduce an outlier detection approach that considers the geospatial consistency between neighboring IP decay curves. However, their approach to data uncertainty estimation relies on computing the deviation of the decay curves from smooth power-law models. In contrast, our VAE approach avoids the power-law assumption and instead uses the entire distribution of $\mathcal{D}_{\textrm{Train}}$ as references curves (see Figure~\ref{fig:data-distribution-zdim=2}). In addition, we obtain the genuine data uncertainty by analyzing $p_{\theta}({\mathbf{x}\given \mathbf{z}})$, allowing the evaluation of any denoised data quantile (see Figures~\ref{fig:2496},~\ref{fig:field-examples-zdim=2-781974-484961-572179}~and~\ref{fig:comparison-field-denoise-28}). Finally, the IP-VAE does not consider geospatial information and was designed to be applicable anywhere.

More recently, \cite{barfod_automatic_2021} described the application of artificial neural networks to automate part of the IP decay curve processing workflow. Reduced processing times and enhanced processing consistency are the primary motivations for their contribution. However, the approach presented therein relies on supervised learning, which required manual labeling of several thousands of IP measurements to produce predictions subsequently. Labeling IP data is not as straightforward as labeling pictures for supervised image classification tasks. Consequently, it is possible that the IP interpreter's inherent biases affected data labeling and propagated to the model. Moreover, the neural network described in \cite{barfod_automatic_2021} only integrates a small number of IP data (2 946 to be precise), all compiled from a unique survey. The authors provided no evidence that the trained model can process new surveys from different geolocations, although supervised learning applications for petrophysical data are prone to high local variability (nugget effect) and overfitting to specific geological contexts \citep[see][]{cate_machine_2017, berube_predicting_2018}. 

Furthermore, neural networks trained on small data sets are susceptible to data set shift when applied to broader test data. Specific causes of data set shift include \citep{quinonero-candela_dataset_2009}: simple covariate shift, sample selection bias (such as training with only one IP survey), imbalanced data sets (different IP surveys that have different S/N), and domain shifts (a change in IP measurement protocol or geolocation). In comparison, we trained the IP-VAE using over 1.6M IP measurements from various locations, survey conditions, subsurface lithologies, and electrode configurations to minimize the potential for data set shift. Most importantly, training the IP-VAE required no manual labeling whatsoever. Finally, the approach presented in \cite{barfod_automatic_2021} relies on the Cole-Cole model to detect outliers, whereas the outlier detection properties of the IP-VAE are data-driven.

\subsection{Parametrization of time-domain IP}
The squared Euclidean norm reconstruction term in $\mathcal{L}_\textrm{VAE}$ implies that the IP-VAE aims to learn an arbitrary function that fits IP data in the least-squares sense. This function, which was imposed as a power-law in \cite{flores_orozco_decay_2018}, is here replaced by a latent distribution of reference curves expressed as $q_{\phi}({\mathbf{z}\given \mathbf{x}})$ and learned from $\mathcal{D}_{\textrm{Train}}$. The various dimensions of this distribution can capture various data features (e.g., amplitude, shape, noise). We observed no significant difference in the generative and reconstruction properties of the IP-VAE using latent dimensions of 1, 2, 4, and 6 (Table~\ref{tbl:latent-comparison}). This result indicated that an excellent reconstruction of the entire data space was possible with a one-dimensional latent space. In other words, the 1\,600\,319 IP curves contained in $\mathcal{D}_{\textrm{Train}}$ were compressible to a single real-valued scalar parameter, preserving the decay shape and losing the high-frequency noise. This observation suggests that many empirical and mechanistic IP models suffer from over-parametrization in the time domain. This interpretation is consistent with a remark made in \cite{bucker_electrochemical_2018}, wherein the authors mention that one drawback of mechanistic models is that electrochemical parameters are particularly ill-constrained and that it is difficult to adjust their large number of parameters to fit data. 

The Cole-Cole model, in its simplest form, describes IP data with only three parameters (chargeability, Cole-Cole exponent, and relaxation time) and an additional parameter for the direct current resistivity \citep[see][]{pelton_mineral_1978}. The results of our experiments suggested that a single parameter, closely related to $\bar{m}$, was sufficient to describe time-domain IP data (see Table~\ref{tbl:latent-comparison} and Figure~\ref{fig:2d-latent-space-zdim=2}). Inverting time-domain IP data for the Cole-Cole exponent and relaxation time may thus be ambiguous because these parameters carry limited information. Our observation is similar to one made in \cite{revil_spectral_2014}, wherein the authors suggested that the Cole-Cole exponent could be a superfluous parameter when fitting data because it rarely deviates from a value of 0.5 (a Warburg impedance).

\subsection{Future research}
This study showed a promising future for applications of unsupervised deep learning methods in the field of IP signal processing. We concentrated our efforts on modeling and processing IP in its time-domain and windowed form, as most practical applications measure it. However, it would be interesting to develop VAEs that learn the latent representation of full-waveform IP decay curves. In addition to the powerful denoising feature of VAEs, analysis of the whole waveform latent space would provide further quantitative evidence toward an ideal parametrization of IP data. In other words, it may provide the answer to the following question: Is it worth it to collect full-waveform IP data if it turns out that the average chargeability is the only parameter with meaning that we can extract from it? With the increasing availability of full-waveform IP receivers---and therefore growing data compilations---we will be conducting further research on the use of VAEs to model the IP waveform. Full waveform IP decays can contain hundreds or thousands of sequential voltage samples. With this added dimensionality of the IP time series, more sophisticated classes of neural networks are desirable. For example, Long Short-Term Memory networks \citep{hochreiter_long_1997}, a type of recurrent neural network with feedback connections that can adapt to variable-length input sequences, has interesting forecasting properties for IP data. The forecasting properties of recurrent neural networks may also be the key to artificially extending time-domain IP data. In particular, forecasting the pure IP effect at shorter times may assist in modeling and eliminating electromagnetic coupling \citep{dahlin_improvement_2012}.

Another avenue for future research is applying unsupervised deep learning methods to frequency-domain (spectral) IP data. This data consists of complex electrical resistivity---or conductivity---measurements in the MHz to kHz range. Experiments with VAE and complex resistivity data may reveal interesting insights into recently proposed mechanistic models \citep[e.g.,][]{revil_induced_2015, misra_interfacial_2016, bucker_electrochemical_2018}. In particular, further work can provide answers to the following research questions: 
\begin{itemize}
    \item Are all mechanistic model parameters interpretable as latent data representations?
    \item Can VAEs denoise low-S/N spectral IP measurements (e.g., in hydrogeological applications where metallic particles are not present to amplify the IP signal)?
    \item Can VAEs learn to verify Kramers--Kronig relations for increasingly complex mathematical models? In other words, can the real part of complex resistivity be encoded and decoded to reconstruct its corresponding imaginary part?
\end{itemize}
Answering these questions will be challenging because the data collection process for spectral IP is much slower than it is for time-domain IP. For example, a total frequency sweep using the SIP-Fuchs III instrument takes approximately 90 minutes, whereas a time-domain IP measurement with the ELREC Pro receiver requires only a few seconds. Consequently, spectral IP is not as commonly used for field applications and is more adapted to studying small amounts of samples in the laboratory \citep[e.g.,][]{berube_mineralogical_2018}. Experiments with VAEs and spectral IP may only be possible using synthetic data for now.

\section{Conclusions}
We trained a deep VAE to encode an extensive compilation of 1\,600\,319 IP measurements collected in various regions of Canada, the United States, and Kazakhstan. Doing so, we made two contributions to the practical and fundamental aspects of the induced polarization method. Our practical contribution is releasing an open-source Python package enabling users to process IP data with a pre-trained implementation of the IP-VAE. We demonstrated four applications of this contribution:
\begin{enumerate}
    \item Generating representative synthetic data.
    \item Denoising IP measurements, including data uncertainty estimation.
    \item Evaluating and interpreting survey S/N.
    \item Detecting outliers using the RMSE reconstruction error.
\end{enumerate}
Our contribution on the fundamental aspect is determining a new bound on the number of parameters required to describe time-domain IP data. We demonstrated that one-dimensional latent vectors (a single, real-valued scalar latent variable) could capture most of the meaningful information in IP decays. Furthermore, this scalar variable carried the meaning of the average chargeability. These results suggested that the shape of the IP decays in our extensive compilation were, in fact, relatively similar, with the average chargeability being the main discriminant between them. Based on this evidence, we concluded that inverse modeling of time-domain IP data with any empirical (e.g., Cole-Cole) or mechanistic model governed by more than one free parameter is ambiguous. Traditional inverse modeling using the average chargeability is thus entirely justified.

\section{Acknowledgments}
C. L. Bérubé's new faculty start-up grant funded this research (Programme PIED 2021, Polytechnique Montréal). We are grateful to Abitibi Geophysics Inc for providing the IP survey compilation and thank A. A. Forestier for help with data processing. 

\newpage
\section{Symbols and nomenclature}

\begin{longtable*}{lll} 
    IP &=& induced polarization\\
    VAE &=& variational autoencoder\\
    $V_\textrm{p}$ &=& the primary voltage\\
    $V_\textrm{s}$ &=& the secondary voltage\\
    $m$ &=& the chargeability IP parameter\\
    $\rho_0$ &=& the subsurface direct current resistivity\\
    $\rho_\infty$ &=& the subsurface resistivity at infinitely high frequency\\
    $t_j$ &=& the duration of a chargeability integration window, $j = 1, 2, \dots, d$\\
    $\bar{m}$ &=& chargeability averaged over $d$ integration windows\\
    $d$ &=& dimensionality of the IP decays, $d=20$ in this study\\ 
    $\mathcal{D}_{\textrm{Train}}$ &=& training data set, $\mathcal{D}_{\textrm{Train}} = [\mathbf{x}^{(1)}, \mathbf{x}^{(2)}, \dots, \mathbf{x}^{(n)}]$, where $n=1\,600\,319$\\
    $\mathcal{D}_{\textrm{STest}}$ &=& synthetic test data set generated by the IP-VAE decoder\\
    $\mathcal{D}_{\textrm{FTest}}$ &=& field test data set containing high and low S/N examples\\
    AE &=& autoencoder\\
    $\phi, \theta$ &=& trainable parameters of the encoder and decoder, respectively\\
    $g_{\phi}$ &=& a deterministic encoder\\
    $\mathbf{x}$ &=& an IP decay in the input data space, $\mathbf{x} = (x_1, x_2, \dots, x_d)$\\
    $\mathbf{z}$ &=& a vector in the autoencoder latent space, $\mathbf{z} = (z_1, z_2, \dots, z_K)$\\
    $f_{\theta}$ &=& a deterministic decoder\\
    $\mathbf{x'}$ &=& a reconstructed IP decay, $\mathbf{x'} = (x'_1, x'_2, \dots, x'_d)$\\
    $\mathcal{L}_{\textrm{AE}}$ &=& the AE loss function\\
    $p_{\theta}({\mathbf{z}\given \mathbf{x}})$ &=& the VAE posterior distribution\\
    $p_{\theta}(\mathbf{z})$ &=& prior distribution of the latent space\\
    $p_{\theta}(\mathbf{x})$ &=& evidence, also known as marginal likelihood\\
    $q_{\phi}({\mathbf{z}\given \mathbf{x}})$ &=& a probabilistic encoder\\
    $p_{\theta}({\mathbf{x}\given \mathbf{z}})$ &=& a probabilistic decoder\\
    $\mathcal{N}(\boldsymbol{\mu}, \boldsymbol{\sigma})$ &=& an isotropic normal distribution parametrized by $\boldsymbol{\mu}$ and $\boldsymbol{\sigma}$\\
    $\boldsymbol{\mu}$ &=& vector means of a normal distribution\\
    $\boldsymbol{\sigma}$ &=& vector standard deviations of an isotropic normal distribution\\
    $\boldsymbol{\epsilon}$ &=& auxiliary random vector for the reparametrization trick, $\boldsymbol{\epsilon}\in\mathcal{N}(0,\mathbf{I})$\\
    $\mathbf{I}$ &=& the identity matrix\\
    $\odot$ &=& the Hadamard product\\
    $D_{\textrm{KL}}$ &=& Kullback–Leibler divergence\\
    NLL &=& negative log-likelihood\\
    $\mathcal{L}_{\textrm{VAE}}$ &=& the variational autoencoder loss function\\  
    $K$ &=& dimensionality of the VAE latent vectors\\
    $h^{(l)}$ &=& the $l$\textsuperscript{th} hidden layer in the VAE\\
    MA &=& a moving average filter\\
    $M$ &=& order of the MA filter\\ 
    EMA &=& an exponential moving average filter\\
    $\alpha$ &=& parameter of the EMA filter, $\alpha\in [0, 1]$\\
    $G(\omega_\textrm{n})$ &=& frequency response of a Butterworth filter\\
    $\omega_\textrm{n}$ &=& normalized cutoff frequency of a Butterworth filter\\
    RMSE &=& root mean square error reconstruction metric\\
    $\sigma_\textrm{noise}$ &=& standard deviation of the Gaussian noise added to $\mathcal{D}_{\textrm{STest}}$
\label{tbl:symbols}
\end{longtable*}

\section{Data and materials availability}
We released a pre-trained PyTorch implementation of the IP-VAE model as an open-source Python package for the geophysics community. The code and some examples are accessible in the following repository: \url{https://doi.org/10.5281/zenodo.5148538}.

\clearpage 
\append{Training data}
\label{s:appA}
\setcounter{table}{0}
\renewcommand{\thetable}{A-\arabic{table}}

\begin{longtable}{lrrrrrrrrr}
\caption{Description of the IP-VAE training data set. In the survey column, the first two digits indicate the year (19: 2019) and the two-letter codes indicate the region (NW: Northwest Territories, MB: Manitoba, NL: Newfoundland-Labrador, NT: Ontario, QC: Québec, FU: Kazakhstan, US: United States). $N$: number of IP decays per survey. $a$: dipole spacing in m. $\widetilde{R}$: median contact resistance in k$\Omega$. $\widetilde{V}_\textrm{p}$: median primary voltage in mV. $\widetilde{I}$: median injected current in mA. $\widetilde{C}$: median empirical confidence score. RMSE: median IP-VAE root-mean-square error in mV/V. S/N: median IP-VAE peak signal-to-noise ratio in dB. $t_\textrm{p}$: survey processing time, in s, required for 100 IP-VAE realizations on the Apple M1 chip.}
\label{tbl:dtrain}\\
\hline
   Survey &   $N$ &   $a$ & $\widetilde{R}$ & $\widetilde{V}_\textrm{p}$ & $\widetilde{I}$ & $\widetilde{C}$ & RMSE &  S/N & $t_\textrm{p}$ \\
\hline \hline
\endfirsthead
\caption[]{(continued)} \\
\hline
   Survey &   $N$ &   $a$ & $\widetilde{R}$ & $\widetilde{V}_\textrm{p}$ & $\widetilde{I}$ & $\widetilde{C}$ & RMSE &  S/N & $t_\textrm{p}$ \\
\hline \hline
\endhead
\hline \hline
\multicolumn{10}{r}{{Continued on next page}} \\
\hline \hline
\endfoot

\hline
\endlastfoot
 18FU058a &  9791 &  6.25 &             0.7 &                          7 &            4700 &            72.3 & 0.78 & 15.8 &            2.1 \\
 18FU088a &  4514 & 18.75 &             4.8 &                         11 &            1600 &            76.1 & 0.82 & 24.4 &            1.0 \\
 18FU088b &  5424 & 18.75 &             4.5 &                         16 &            1650 &            83.9 & 0.65 & 27.7 &            1.2 \\
 18FU088c &  1542 & 18.75 &             7.1 &                         14 &            1320 &            82.8 & 0.59 & 26.9 &            0.3 \\
 18FU088d &  3785 & 18.75 &             9.2 &                         10 &            1050 &            75.6 & 0.83 & 25.4 &            0.8 \\
 18FU088e &  4275 & 18.75 &             7.7 &                          9 &            1190 &            73.5 & 0.91 & 24.4 &            0.9 \\
 18FU088f &  3464 & 18.75 &             6.9 &                         14 &            1300 &            81.3 & 0.65 & 26.0 &            0.7 \\
 18FU089a & 23650 &  6.25 &             2.5 &                          8 &            2400 &            68.7 & 0.97 & 21.8 &            5.0 \\
 18FU089b &  6533 &  6.25 &             2.6 &                          9 &            2500 &            70.3 & 0.89 & 22.1 &            1.4 \\
 18MB073a &  6742 & 12.50 &           193.1 &                         85 &             120 &            80.0 & 1.00 & 26.0 &            1.4 \\
 18MB073b &  3280 & 12.50 &             1.0 &                        189 &            1450 &            91.0 & 0.53 & 21.5 &            0.7 \\
 18MB073c &  4049 & 12.50 &             4.7 &                        325 &             900 &            97.8 & 0.28 & 32.2 &            0.9 \\
 18NL060a & 12907 & 25.00 &             2.9 &                        792 &             400 &            64.4 & 9.62 & 17.6 &            2.7 \\
 18NL090a &  6090 & 18.75 &             1.2 &                        158 &            1000 &            99.2 & 0.17 & 31.7 &            1.3 \\
 18NT004a & 50374 & 18.75 &            15.1 &                         94 &             270 &            89.7 & 0.47 & 27.9 &           10.7 \\
 18NT021a &  2070 & 12.50 &             0.2 &                         25 &            3500 &            83.8 & 0.46 & 16.7 &            0.4 \\
 18NT021b &   860 & 12.50 &             0.3 &                         27 &            2500 &            81.6 & 0.49 & 16.3 &            0.2 \\
 18NT047a & 10111 & 12.50 &             8.6 &                        466 &             280 &            97.9 & 0.34 & 29.2 &            2.1 \\
 18NT054a & 38091 & 12.50 &             8.3 &                        116 &             770 &            96.2 & 0.27 & 27.2 &            8.1 \\
 18NT054b & 56535 & 12.50 &             1.6 &                        113 &            1100 &            96.9 & 0.30 & 29.4 &           12.0 \\
 18NT070a & 12968 & 12.50 &             0.5 &                         74 &            2170 &            97.3 & 0.33 & 20.1 &            2.7 \\
 18NW040a &  1544 & 12.50 &            32.4 &                        241 &             890 &            91.6 & 0.37 & 29.0 &            0.3 \\
 18NW040b &  7971 & 12.50 &            30.7 &                          6 &            1100 &            61.1 & 3.46 & 14.1 &            1.8 \\
 18NW040c &  3805 & 12.50 &            33.5 &                        130 &              80 &            87.2 & 0.67 & 30.5 &            0.8 \\
 18QC001a & 44011 & 18.75 &             0.6 &                        115 &            1900 &            99.3 & 0.25 & 28.8 &            9.2 \\
 18QC002a & 10672 & 25.00 &             0.7 &                         60 &            2100 &            94.1 & 0.33 & 19.3 &            2.3 \\
 18QC002b &  6670 & 25.00 &             0.8 &                        146 &            2000 &            99.4 & 0.30 & 18.5 &            1.4 \\
 18QC016a &  1185 & 25.00 &             6.5 &                         40 &            5000 &            74.2 & 0.82 & 19.6 &            0.3 \\
 18QC018a & 16843 & 12.50 &             8.5 &                        187 &             230 &            97.9 & 0.34 & 30.0 &            3.6 \\
 18QC020a & 22051 & 12.50 &             0.9 &                        198 &            1060 &            99.2 & 0.25 & 33.1 &            4.7 \\
 18QC029a &  6404 &  6.25 &             4.0 &                        243 &             500 &            99.2 & 0.35 & 34.2 &            1.3 \\
 18QC029b &  2360 & 18.75 &             2.1 &                        483 &             775 &            99.9 & 0.26 & 33.7 &            0.5 \\
 18QC031a & 90146 & 12.50 &             3.8 &                        245 &             810 &            98.4 & 0.28 & 27.3 &           19.5 \\
 18QC033a &  3075 & 12.50 &             1.4 &                        171 &            1700 &            96.8 & 0.34 & 26.9 &            0.7 \\
 18QC037a &  2059 & 25.00 &             4.8 &                         63 &            1200 &            97.2 & 0.27 & 29.6 &            0.4 \\
 18QC038a &  2040 & 18.75 &             0.8 &                         49 &            1700 &            99.1 & 0.53 &  4.3 &            0.4 \\
 18QC038b &  1700 & 18.75 &             0.8 &                         26 &            1600 &            87.3 & 0.50 & 20.3 &            0.4 \\
 18QC038c &  2040 & 18.75 &             0.9 &                         58 &            1550 &            98.5 & 0.41 & 13.6 &            0.4 \\
 18QC039a &  3028 & 18.75 &             0.3 &                         81 &            2500 &            88.6 & 0.52 & 19.6 &            0.6 \\
 18QC048a &  4199 & 12.50 &             6.0 &                        126 &             570 &            94.4 & 0.43 & 28.5 &            0.9 \\
 18QC049a & 42888 & 18.75 &             8.3 &                        219 &             500 &            92.1 & 0.43 & 28.1 &            9.1 \\
 18QC056a &  4207 &  6.25 &             0.7 &                        228 &            1920 &            98.7 & 0.28 & 21.1 &            0.9 \\
 18QC057a &  3799 &  6.25 &             2.3 &                        370 &            1050 &            97.1 & 0.33 & 29.5 &            0.8 \\
 18QC069a &  6132 & 18.75 &             1.7 &                        102 &            1500 &            97.6 & 0.43 & 13.0 &            1.3 \\
 18QC076a & 60054 &  6.25 &             8.4 &                        148 &             400 &            91.4 & 0.43 & 32.0 &           12.7 \\
 18US026a &  6445 & 25.00 &             8.3 &                         60 &             770 &            95.8 & 0.45 & 32.3 &            1.3 \\
 18US051a & 14270 & 82.50 &             2.1 &                         21 &            1300 &            55.6 & 2.98 & 14.4 &            3.0 \\
 19FU020a & 20280 &  6.25 &             0.5 &                         14 &            3600 &            90.3 & 0.65 & 29.4 &            4.2 \\
 19FU033a & 16821 & 18.75 &             2.0 &                         23 &            3000 &            80.8 & 0.61 & 20.3 &            3.5 \\
 19FU046a & 58949 &  6.25 &             2.5 &                         11 &            2600 &            74.0 & 0.78 & 24.8 &           12.3 \\
 19MB002a & 14304 & 12.50 &             0.6 &                        130 &            2150 &            86.9 & 0.52 & 17.9 &            3.0 \\
 19MB009a &  6914 & 25.00 &            46.1 &                         56 &            1650 &            71.1 & 1.11 & 20.3 &            1.5 \\
 19NT031a &  8910 & 18.75 &             7.9 &                        247 &             430 &            94.1 & 0.37 & 26.9 &            1.9 \\
 19NT031b & 10770 &  6.25 &             2.4 &                        178 &            1000 &            94.4 & 0.33 & 24.6 &            2.2 \\
 19NT054a & 12878 & 12.50 &             3.1 &                         48 &             870 &            94.0 & 0.30 & 24.7 &            2.7 \\
 19NW019a &   262 & 12.50 &           287.3 &                        141 &              10 &            86.3 & 0.66 & 28.5 &            0.1 \\
 19NW019b &  1965 & 12.50 &             0.4 &                         67 &              35 &            74.3 & 0.75 & 15.7 &            0.4 \\
 19NW019c &  3174 & 12.50 &           309.1 &                        136 &              24 &            82.0 & 0.79 & 26.3 &            0.7 \\
 19NW019d &  4833 & 12.50 &            47.2 &                         22 &             410 &            79.7 & 4.56 & 17.7 &            1.0 \\
 19QC003a & 98290 & 18.75 &            12.7 &                        118 &             320 &            91.3 & 0.40 & 28.2 &           21.1 \\
 19QC005a &  7736 & 12.50 &             0.3 &                        149 &            2000 &            87.7 & 0.66 &  7.7 &            1.7 \\
 19QC006a & 12048 & 12.50 &             0.7 &                         96 &            1900 &            99.5 & 0.33 & 30.9 &            2.5 \\
 19QC016a & 13810 & 25.00 &             0.6 &                         57 &            2400 &            93.1 & 0.52 &  9.5 &            2.9 \\
 19QC021a &  1578 & 12.50 &             1.9 &                        502 &             280 &            97.1 & 0.29 & 28.8 &            0.3 \\
 19QC021b &  8597 & 12.50 &            16.5 &                        191 &             260 &            87.7 & 0.38 & 28.2 &            1.8 \\
 19QC021c & 12015 & 12.50 &             4.9 &                        322 &             360 &            95.7 & 0.38 & 30.3 &            2.5 \\
 19QC032a & 20069 & 12.50 &             3.4 &                        254 &            1050 &            85.8 & 0.47 & 18.9 &            4.2 \\
 19QC032b & 17311 & 18.75 &             1.3 &                        140 &            1450 &            86.9 & 0.44 & 17.1 &            3.6 \\
 19QC047a & 14652 & 12.50 &             0.6 &                        257 &            1000 &            96.5 & 0.27 & 26.3 &            3.0 \\
 19QC047b & 27828 & 12.50 &             3.1 &                        209 &             700 &            96.7 & 0.25 & 27.7 &            5.8 \\
 19QC047c & 21743 & 12.50 &             7.7 &                        215 &             400 &            96.9 & 0.26 & 29.0 &            4.5 \\
 19QC063a & 13359 & 12.50 &             3.8 &                        327 &             430 &            99.3 & 0.19 & 31.2 &            2.8 \\
 20FU028a & 14073 &  6.25 &             2.1 &                         20 &            3700 &            82.8 & 0.59 & 25.7 &            3.0 \\
 20FU033a & 14008 & 18.75 &             1.0 &                         32 &            4200 &            81.9 & 0.46 & 24.3 &            3.0 \\
 20FU035a & 14834 &  6.25 &             0.4 &                         10 &            4000 &            73.6 & 0.81 & 22.9 &            3.2 \\
 20FU042a &  3007 & 18.75 &             1.4 &                          8 &            2800 &            61.5 & 1.20 & 16.9 &            0.6 \\
 20FU043a & 88097 &  6.25 &             1.5 &                         14 &            3000 &            80.2 & 0.73 & 24.4 &           18.7 \\
 20FU054a & 26114 &  6.25 &             2.7 &                         22 &            1700 &            84.2 & 0.57 & 25.9 &            5.5 \\
 20FU061a & 18432 &  6.25 &             2.5 &                         19 &            3300 &            80.3 & 0.62 & 25.2 &            3.9 \\
 20FU072a & 20801 & 18.75 &             3.4 &                          6 &            1700 &            70.3 & 1.39 & 21.8 &            4.4 \\
 20FU082a &  3029 & 18.75 &             4.9 &                         38 &            1400 &            90.8 & 0.45 & 27.5 &            0.6 \\
 20NL057a & 18072 & 12.50 &             3.3 &                        121 &             560 &            95.0 & 0.26 & 28.8 &            3.8 \\
 20NL070a &  3085 & 12.50 &             2.5 &                        277 &             650 &            99.2 & 0.19 & 32.0 &            0.7 \\
 20NT004a & 25682 & 12.50 &             9.1 &                        143 &             540 &            96.4 & 0.41 & 32.5 &            5.3 \\
 20NT009a &  6088 & 12.50 &             2.5 &                        403 &            1200 &            99.9 & 0.37 & 17.1 &            1.3 \\
 20NT024a &  2035 & 12.50 &             0.3 &                        319 &            1570 &            99.1 & 0.19 & 27.5 &            0.4 \\
 20NT024b &  6166 & 12.50 &             1.0 &                        424 &            1800 &            96.7 & 0.31 & 25.8 &            1.3 \\
 20NT024c &  3219 & 12.50 &             1.3 &                        436 &            1525 &            99.4 & 0.31 & 31.3 &            0.7 \\
 20NT047a &  8190 &  0.50 &             8.4 &                        316 &             651 &            97.9 & 0.46 & 34.2 &            1.8 \\
 20NT047b &  2698 &  1.00 &             5.9 &                       1330 &            1317 &            96.1 & 0.37 & 29.9 &            0.6 \\
 20NT047c & 16567 & 18.75 &             7.5 &                        240 &             430 &            93.7 & 0.94 & 28.5 &            3.4 \\
 20NT063a & 13550 & 12.50 &             4.9 &                         36 &            1160 &            91.5 & 0.40 & 16.7 &            2.9 \\
 20NT067a & 13633 & 12.50 &             7.8 &                        203 &             300 &            95.8 & 0.29 & 28.4 &            2.9 \\
 20QC001a & 18164 & 12.50 &             0.8 &                        104 &            1500 &            99.4 & 0.35 & 31.5 &            3.8 \\
 20QC007a & 10652 &  6.25 &             0.6 &                        117 &            2350 &            99.9 & 0.36 & 14.4 &            2.3 \\
 20QC008a &  3000 & 12.50 &             3.8 &                        304 &             650 &            99.0 & 0.18 & 30.5 &            0.7 \\
 20QC017a &  5235 & 12.50 &             0.6 &                        414 &            1950 &           100.0 & 0.22 & 23.1 &            1.1 \\
 20QC020a &   956 & 25.00 &            15.9 &                         37 &            4000 &            69.0 & 0.73 & 20.7 &            0.2 \\
 20QC020b &  5110 & 12.50 &             6.0 &                         78 &             380 &            71.9 & 0.75 & 21.6 &            1.1 \\
 20QC040a &  4969 & 18.75 &             0.8 &                        147 &            2050 &            96.7 & 0.45 &  9.7 &            1.1 \\
 20QC041a & 23629 & 18.75 &             1.2 &                        333 &            2000 &            89.9 & 0.37 & 25.4 &            5.0 \\
 20QC048a & 17760 & 18.75 &             0.7 &                         95 &            2100 &            99.2 & 0.26 & 22.6 &            3.8 \\
 20QC051a & 81852 &  6.25 &             0.9 &                        243 &            1900 &            95.8 & 0.37 & 17.7 &           17.4 \\
 20QC074a & 11572 & 18.75 &             1.6 &                        119 &            1000 &            91.3 & 0.35 & 24.4 &            2.4 \\
 20US044a &   687 & 50.00 &             9.9 &                        220 &            4000 &            99.2 & 0.33 & 31.9 &            0.2 \\
 20US044b &   186 & 25.00 &            13.9 &                        230 &            3550 &            99.3 & 0.29 & 35.9 &            0.0 \\
 21FU013a &  1574 & 18.75 &            13.4 &                        107 &            1300 &            88.8 & 0.47 & 26.8 &            0.4 \\
 21FU013b &  3012 & 18.75 &             5.4 &                         25 &            1200 &            85.4 & 0.33 & 23.9 &            0.6 \\
 21NT011a & 17438 & 50.00 &            16.4 &                        123 &             602 &            95.6 & 0.51 & 30.5 &            3.7 \\
 21QC005a &  7394 & 18.75 &             1.0 &                        438 &             850 &            99.1 & 0.30 & 25.3 &            1.5 \\
\end{longtable}

\setcounter{table}{0}
\renewcommand{\thetable}{\arabic{table}}

\bibliographystyle{seg}  
\bibliography{references}

\end{document}